\documentclass[preprint,12pt]{elsarticle}

\newtheorem{myDef}{Definition}
\newtheorem{myExa}{Example}

\usepackage{amssymb}
\usepackage{mathtools}
\usepackage{amsfonts,amssymb}
\usepackage{amsmath}
\usepackage{graphicx}
\usepackage{subcaption}
\usepackage[noend]{algpseudocode}
\usepackage{algorithm,algpseudocode}
\usepackage{multirow}
\usepackage{booktabs}
\usepackage{soul}
\soulregister\cite7
\soulregister\citep7 
\soulregister\citet7
\soulregister\ref7 
\soulregister\pageref7

\journal{Pattern Recognition}

\begin{document}

\begin{frontmatter}



\title{Estimating Fund-Raising Performance for Start-up Projects from a Market Graph Perspective}



\author[1]{Likang Wu}
\ead{wulk@mail.ustc.edu.cn}
\author[1]{Zhi Li}
\ead{zhili03@mail.ustc.edu.cn}
\author[2]{Hongke Zhao}
\ead{hongke@tju.edu.cn}
\author[1]{Qi Liu}
\ead{qiliuql@ustc.edu.cn}
\author[1]{Enhong Chen\corref{cor1}}
\ead{cheneh@ustc.edu.cn}

\cortext[cor1]{Corresponding author}

\address[1]{Anhui Province Key Laboratory of Big Data Analysis and Application, University of Science and Technology of China, No.96, JinZhai Road, Baohe District, Hefei, Anhui, 230026, P.R.China}
\address[2]{College of Management and Economics, Tianjin University, No.92, Weijin Road, Nankai District, Tianjin, 300072, P.R.China}

\begin{abstract}

In the online innovation market, the fund-raising performance of the start-up project is a concerning issue for creators, investors and platforms. Unfortunately, existing studies always focus on modeling the fund-raising process after the publishment of a project but the predicting of a project attraction in the market before setting up is largely unexploited. 
Usually, this prediction is always with great challenges to making a comprehensive understanding of both the start-up project and market environment.
To that end, in this paper, we present a focused study on this important problem from a market graph perspective. Specifically, we propose a Graph-based Market Environment (GME) model for predicting the  fund-raising performance of the unpublished project by exploiting the market environment. In addition, we discriminatively model the \textit{project competitiveness} and \textit{market preferences} by designing two graph-based neural network architectures and incorporating them into a joint optimization stage. Furthermore, to explore the information propagation problem with dynamic environment in a large-scale market graph, we extend the GME model with parallelizing competitiveness quantification and hierarchical propagation algorithm. Finally, we conduct extensive experiments on real-world data. The experimental results clearly demonstrate the effectiveness of our proposed model.

\end{abstract}



\begin{keyword}


Crowdfunding \sep Market Environment Modeling \sep Graph Neural Network
\end{keyword}

\end{frontmatter}


\section{Introduction}
Recent years have witnessed the enormous success of online fund-raising platforms, which aim at helping start-up projects to solicit fundings from the public. These platforms brought new vitality to the innovation market, i.e., making it easy to demonstrate and accomplish innovative ideas of start-up entrepreneurs or individuals across the globe.

When launching a project on online fund-raising platforms, such as Indieg-ogo$\footnote{https://www.indiegogo.com/}$ and Kickstarter$\footnote{https://www.kickstarter.com/}$, creators will carefully illustrate their project backgrounds, reward options and set a target amount of the solicited money. If the money solicited by the public reaches the target amount, the project is successful. However, according to  statistics, only 40\% of the projects succeed in reaching their goals~\cite{liu2017enhancing}. Along this line, the fund-raising performance of the project is a concerning issue for both creators and platforms.

In the literature, there are many studies from both the project view~\citep{li2016project,jin2019estimating} and the market view~\citep{zhao2017tracking, cheng2019success}. These existing researches mainly focus on modeling the fund-raising process after the project published. However, for creators and platform operators, they both want to understand and estimate the fund-raising ability before the project sets up. On the one hand, a good estimation of the fund-raising ability can help creators modify the project designs to cater to current market preferences. On the other hand, since launching an innovation project requires a lot of labor and economic costs, it is important to choose a suitable time to launch the project on market based on the accurate estimation of the project's fund-raising ability. Different from the prior work on tracking the fund-raising process after the project published, this estimation task is mainly based on limited information about the target project without the investors' behaviors. It becomes a non-trivial problem to make an accurate estimation of the fund-raising performance before the project sets up.

In fact, the project's fund-raising performance is always influenced by the market environment, such as the project's competitiveness in the market and the evolution of market preferences. However, it is pretty challenging to model the complex and dynamic market environment to predict fund-raising performance before the target project sets up. Firstly, when a start-up project is launched, it will face fierce competition from other projects in the market. So it is difficult to model the complex competition relationships and predict the competitiveness of the target project. Secondly, market preferences can also have a huge impact on the fund-raising process. As a consequence, how to track the dynamic market preferences is another challenge for our task. Thirdly, with a large number of projects in the market, it is of great challenge to effectively model the market environment.

To conquer the challenges mentioned above, we provide a preliminary study~\cite{DBLP:conf/aaai/WuLZPLC20} on making a prediction on the fund-raising performances of the start-up projects with a graph-based model. To deeply explore the evolution of market environments, we propose a Graph-based Market Environment (GME) model, to explore the market environment of online fund-raising platforms and predict the performance of the target project. Specifically, GME consists of two components, i.e., Project Competitiveness Modeling (PCM) and Market Evolution Tracking (MET), to model the competitive influences of other projects in the market and track the evolution of market preferences, respectively. In the PCM module, we apply a hybrid Graph Neural Network (GNN) to capture the competitiveness pressure of the target project by aggregating competitive influences of other rivals in the market. Additionally, in the MET module, we design a propagation structure to track the market evolution and apply Gated Graph Neural Networks (GG-NNs)~\citep{li2015gated} to model the evolution process. Next, we optimize GME with a joint loss to learn the fund-raising performance of competitive projects and predict the performance of the target project, simultaneously.

Furthermore, we extend the existing work and deeply explore the information propagation issue with market evolution on a large-scale environment graph. In the previous project competitiveness model, Long Short-Term Memory networks (LSTM)~\cite{hochreiter1997long} was used to predict the funding status of each project in the future market. Considering the LSTM-modeled states undermine the possibility of model parallelizing, we firstly propose a novel feedforward competitiveness quantification with prior knowledge to present the future statuses of existing projects in the market. Then, to further explore the information propagation process in the market evolution, we design a hierarchical updating algorithm to address the redundant information problem in the previous MET module. In summary, the contributions of this paper can be summarized as follows:
\begin{itemize}
 \item We conduct a focused study on the environment evolution of the online innovation market and address the fund-raising performance prediction before the project sets up from an environment graph perspective.
 \item To model the market environment, we introduce a dual-aspect GNN framework GME which models the market environment from two aspects, i.e., \textit{project competitiveness} and \textit{market preferences}. Furthermore, to efficiently address the information propagation problem in the large-scale environment graph, we propose the novel feedforward competitiveness quantification and hierarchical updating algorithm to deeply explore the market environment for start-up projects.
 \item We conduct sufficient experiments on real-world datasets of different scales and experimental results clearly demonstrate the effectiveness of our proposed method as well as the superior exploration on the market environment in our GME framework.
 
 
\end{itemize}

The rest of this paper is organized as follows. We will introduce the related work at length in Section 2. The whole framework and technical details are described in Section 3.  Section 4 shows the experimental results of our model and all baselines. Finally, in Section 5, we conclude this work and give the expectation in the future.
\label{Introduction}

\section{Related Work}
In our work, the crowdfunding market environment would be modeled by graph neural networks. Therefore, we review the research from two categories, i.e., online crowdfunding and graph neural networks. 

\subsection{Online Crowdfunding}
Online crowdfunding is a special form of P2P lending, according to the different perspectives of research, it is natural to classify the existing work of online crowdfunding into two directions: Research for individual projects or the market. For the former, the most popular point is the prediction algorithm of success rate. Specifically, ~\citet{li2016project} used the survival analysis theory to model the prediction task and studied the distribution of success times in crowdfunding data, which showed logistic distribution was suitable for learning this problem. A deep learning framework for predicting success probability based on time series was proposed by constructing a sequence-to-sequence model~\cite{jin2019estimating}. To fully utilize multimodal data,  many deep learning models~\cite{cheng2019success,kaminski2019predicting,wang2020prediction,wang2020crowdfunding} that incorporated information from various sources (e.g., visual, textual, and so on) to promote the success rate prediction accuracy. Besides, there were several researchers who paid attention to the time series analysis~\citep{zhao2017tracking, cheng2019success}. For instance, ~\citet{zhao2017tracking,zhao2019voice} tracked the dynamic changes within the fund-raising process to deal with various related tasks. Moreover, in an interesting field, it is necessary to optimize the production supply to help funders design more reasonable perks~\cite{liu2017enhancing}. For the market states modeling, ~\citet{lin2018modeling} proposed a probabilistic generative model to indicate the importance of dynamic market competition, some researchers focused on the recommender systems in the crowdfunding market~\citep{zhang2019personalized, rakesh2015project}, and~\citet{janku2018successful} supported that a good performance led to a strong attraction for the market resource.

Actually, there is a common point that the above methods try to find the pattern of investors' attention or market reaction to published projects. However, we still lack a sufficiently deep and quantitative exploration of the potential performance before the project sets up. There is a very recent work focused on how to predict the fund-raising performances of projects in the crowdfunding market~\citep{DBLP:conf/aaai/WuLZPLC20}. However, in~\citep{DBLP:conf/aaai/WuLZPLC20}, the LSTM-modeled states undermine the possibility of model parallelizing, and the redundant information of propagation tree would depress the market evolution modeling. Here we propose the novel MLP-based competitiveness quantification with prior knowledge and hierarchical updating algorithm to address these two issues respectively. Meanwhile, sufficient corresponding experiments are evaluated in the experimental section.

\subsection{Graph Neural Network}
Graph Neural Network (GNN)~\citep{gori2005new,scarselli2008graph}, especially graph convolutional networks~\citep{bruna2013spectral,henaff2015deep} has shown powerful ability to tackle the network structures due to its theoretical elegance~\citep{bronstein2017geometric}. Extending neural networks to work with graph-structured data was first proposed by~\citet{gori2005new}. With the proposal of spectral method~\citep{bruna2013spectral} and its developments \citep{defferrard2016convolutional,kipf2016semi}, this research area began to attract more and more attention and demonstrated vigorous vitality. In particular, the polynomial spectral filters~\citep{defferrard2016convolutional} reduced the computational cost. Then~\citet{kipf2016semi} proposed the linear filter to make a further simplification. Along with spectral graph convolution, many researchers directly performed graph convolution in the spatial domain~\cite{duvenaud2015convolutional,hamilton2017inductive,chen2020graph}. \citet{bai2020learning} subsequently proposed the BASGCN that bridges the theoretical gap between traditional Convolutional Neural Network (CNN) models and spatially-based GCN models. In addition, there were other representative types of neural networks for graphs. For instance, ~\citet{zhou2007learning} developed the approach for hypergraph embedding and transductive classification, \citet{zhang2019quantum} proposed a quantum-based subgraph convolutional neural network to capture the global topological structure and the local connectivity structure.

GNN has been implemented into several important applications. For recommendation systems, many works utilized graph neural networks to capture the interest similarities among quite a number of customers or the content coherences of every item pair~\citep{DBLP:conf/aaai/MaMZSLC20,DBLP:conf/www/GeWWQH20, chai2019loan}. For social network modeling, the community event discovery and influence propagation were learned by designed GNNs \citep{DBLP:conf/aaai/WuLXWC20, wang2019mcne,wu2021learning}. For P2P lending, the loan requirements and investing lenders can also be matched by hybrid GCN~\citep{chai2019loan}.
Compared with traditional network embedding methods, the end-to-end GNN-based models were more suitable for supporting the platform's running~\citep{wang2019mcne}. 

Most of the above algorithms were applied to existing networks such as social network, items relation network, and so on, but there were also virtual graphs created by researchers according to actual problem scenario. For instance, ~\citet{feng2019temporal} created the relations between every pair (two industries) based on a knowledge base to predict stock trends. Graph kernels were also able to play an important role in the financial time series analysis by associating it with the classical dynamic time warping framework~\cite{bai2020entropic,bai2019quantum}. And, in the multimedia field, \citet{cucurull2019context} promoted the fashion compatibility prediction problem using a graph convolutional network that learned to generate product embeddings conditioned on context. Similarly, we deeply analyzed the project competition phenomenon and market evolution process to create corresponding graphs and proposed a novel graph neural network framework to capture the characters of the market environment.

\section{Framework}
In this section,  we first define our research problem formally and present the existing data features consist of static features and market dynamic features. Next, along a market modeling line, we propose a Graph-based Market Environment model (GME) to tackle our task. We will describe the model from two aspects of modeling the market environment, i.e., project competitiveness modeling and market evolution tracking.

\subsection{Preliminaries}
As the main task is to estimate the fund-raising performance before the project sets up, we first need to define what is the project fund-raising performance. Actually, for a start-up project on the innovation market, it is not proper to just use the fund-raising amounts to measure corresponding performance. The reason is that the same amount of funding in the raising process may mean different performance for the projects with various pledged goals. For example, an electric vehicle project with a goal of 1 million dollars may raise 1,000 dollars in a day, at the same time, a book-writing project with the goal of 800 dollars may raise 100 dollars in a day. Although electric vehicle raises a much huger amount of money than the book-writing project, the book-writing is obviously easier to accomplish the fund-raising goal. So in this work, we use the degree of completion of pledged goal to measure the target value. In detail, for target project $i$, we design the quantitative definition of the fund-raising performance $y_i$ as follow:

\begin{figure}[t]	
	\centering
	\begin{subfigure}[t]{.4\columnwidth}
		\centering
		\includegraphics[width=\columnwidth]{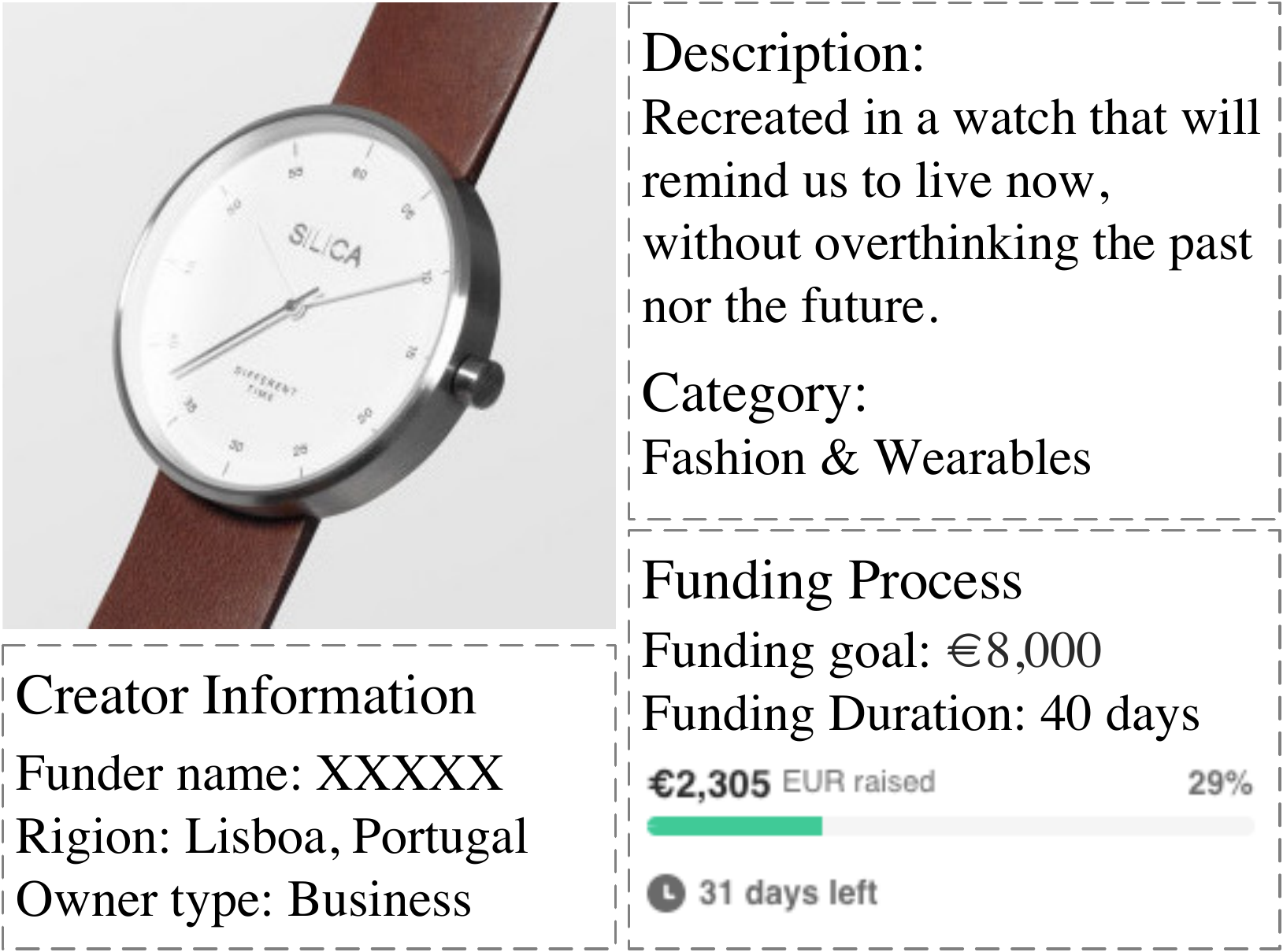}
		\caption{}\label{fig:project}		
	\end{subfigure}
	\quad
	\begin{subfigure}[t]{.55\columnwidth}
		\centering
		\includegraphics[width=\columnwidth]{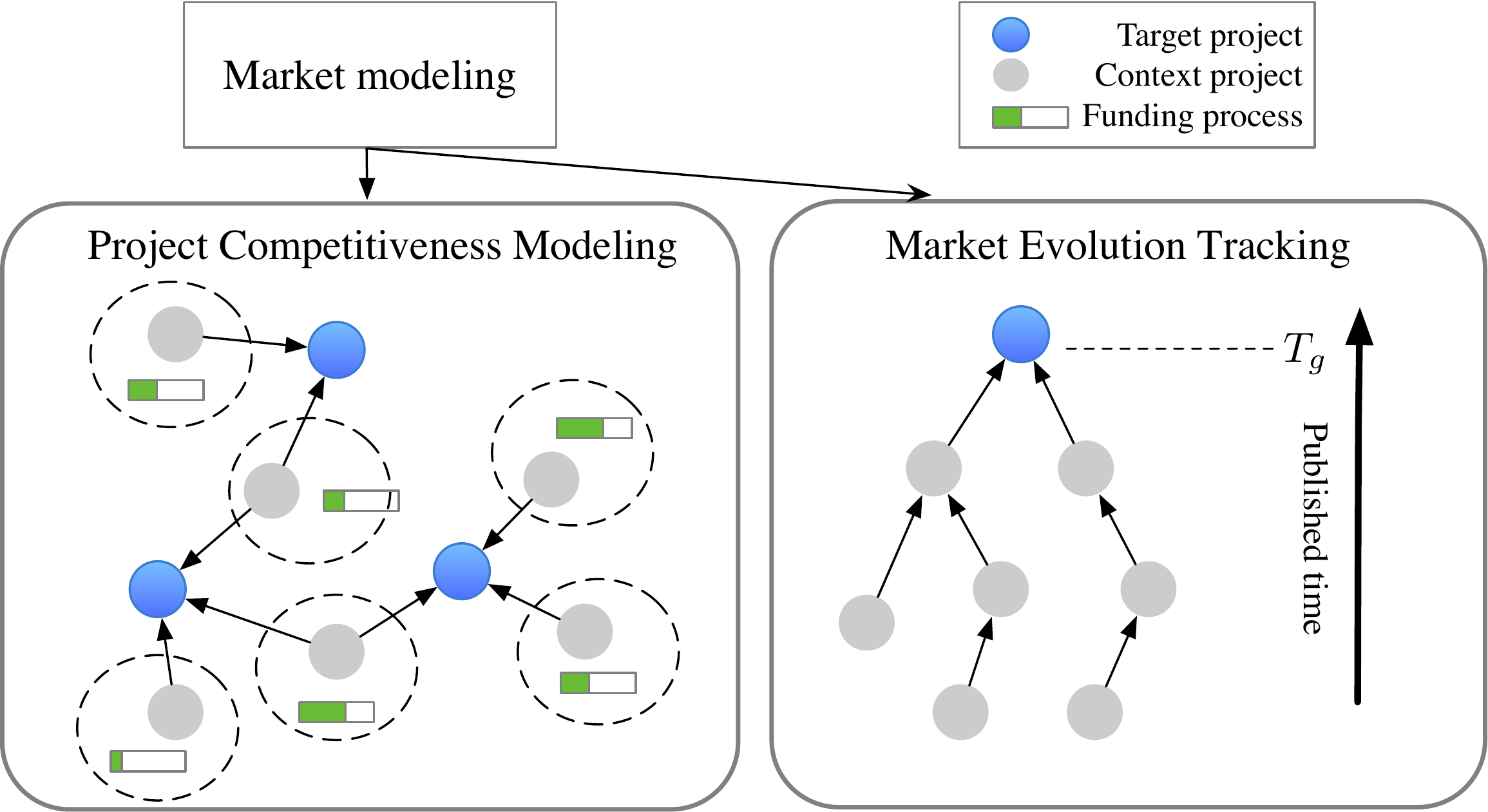}
		\caption{}\label{fig:illusion}
	\end{subfigure}
	\caption{(a) A toy example of a crowdfunding project. (b) The two considerations of market environment modeling. }\label{fig:description}
\end{figure}

\begin{align}
\label{eqy} y_i = log_2\left(1 + \frac{\alpha_i}{g_i}\right),
\end{align}

where we will predict the normalized value $y_i$. The pledged goal of a project is denoted by $g_i$. As our estimation of fund-raising performance is before the project sets up, we just use funding amount $\alpha_i$ within the first $\mathcal{T}$ hours (in our experiments, $\mathcal{T} = 24, 48$), which aims to eliminate influences of the project's content changes after setting up. $\frac{\alpha_i}{g_i}$ represents the achievement percentage of declared pledged goal. We use the $log_2(\cdot)$ function to normalize $y_i$, in this way, the large variation between minimum and maximum~\citep{wang2018learning} could be suppressed.

Due to the market environments are different at different start-up time, they would affect the fund-raising performance~\citep{janku2018successful}. As mentioned before, modeling the market environment is crucial for the estimation of a project's fund-raising performance. Here we would consider the static and dynamic features about the market when predicting the project fund-raising performance.
Therefore, we formally define the studied problem as follows:
\begin{myDef}
	\textbf{Fund-raising Performance Estimation}. Given the target project $g$ which plans to launch at time $T_{g}$ and the static features $\mathbf{X}^{T_{g}}$ of all related projects with the dynamic fund-raising sequence $S^{T_{g}}$ in the market before $T_{g}$, the task focuses on predicting the fund-raising performance $y_g$ for target project $g$.
\end{myDef}

\begin{table}[t]
	\caption{The information of static features.}\smallskip
	\centering
	\resizebox{.45\columnwidth}{!}{
		\smallskip\begin{tabular}{ccccccc}
			\hline
			\textbf{Features} & \textbf{Type}\\
			\hline
			Project Description & Text\\
			Project Category & Categorical\\
			Creator Type & Categorical\\
			Currency & Categorical\\
			Declared Funding Duration & Numerical\\
			Declared Pledged Goal & Numerical\\
			\hline
		\end{tabular}
	}
	\label{table:features}
\end{table}
\subsubsection{Feature Description}
We detailedly introduce the static features and market dynamic features mentioned in \textbf{Definition 1}.
For the former, there are a total of $N$ projects, and we concatenate the embedding vectors of static information like project descriptions, project category, and declared funding duration and so on. We show all static features in Table~\ref{table:features} and a toy example of crowdfunding project in Figure~\ref{fig:project}. Specifically, we use one-hot encoding to discretize numerical values, which is able to improve the signal-to-noise ratio~\cite{liu2002discretization}. Fitting numerical features to bins reduces the impact that small fluctuations in the data have on the model, and small fluctuations are often just noises. And, following ~\citet{jin2019estimating}, text features are converted into numerical vectors via doc2vec algorithm~\cite{le2014distributed}. The instance matrix $\mathbf{X} = {\left[\ \mathbf{x}_1, \mathbf{x}_2,...,\mathbf{x}_N \ \right]}^{\mathrm{T}}$ is created by all existing projects and each row denotes the feature of one project.

For market dynamic features, there are also lots of investment records in the crowdfunding market. Although our aim focuses on the target project's risk reduction before the published time, which means we are not able to implement time series based prediction method. It is important to use the historic fund-raising processes of the published projects in the market to reflect the rivals' competitiveness and market evolution. For a fund-raising process of project $i$, we give a fund-raising sequence $S_i = \{ \langle v_1, t_1 \rangle , \langle v_2, t_2 \rangle , ... , \langle  v_{\vert S_i \vert}, t_{\vert S_i \vert} \rangle \}$, where $v$ presents the investment amount of this project at time $t$.

\subsection{Project Competitiveness Modeling}
Due to the limited resource, a project will run in a competitive environment if it has been published. And the core of this section is to explore the way that model the competitiveness pressure on our target projects. Along this line, we firstly give the competitiveness quantification methods consist of LSTM modeling and the more concise feedforward modeling, and the latter has a better fit with large-scale data sources. Secondly, a graph structure is constructed to get the competitiveness aggregation of projects' rivals. The above two steps constitute our Project Competitiveness Modeling (PCM) module which can model one competitive market environment at a certain moment.

\subsubsection{Competitiveness Quantification via LSTM}
\begin{figure*}[t]
\centering
\includegraphics[width=.98\textwidth]{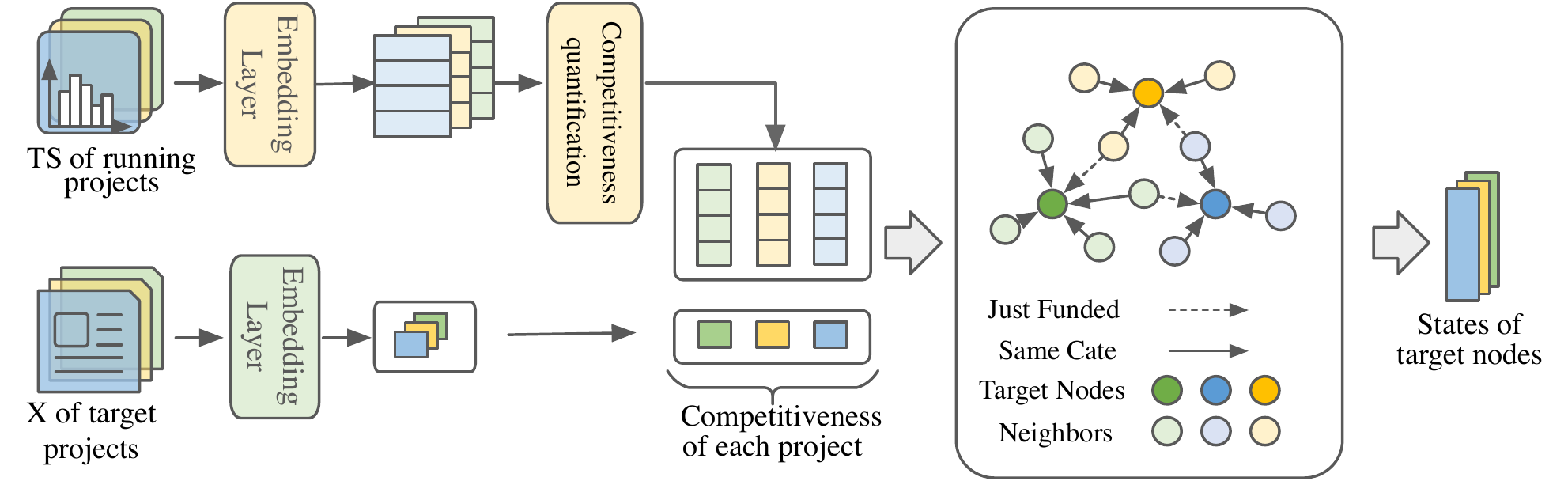} 
\caption{The overview of Project Competitiveness Modeling (PCM) which captures the competitive relationships between the target projects and others.}
\label{fig:PCM}
\end{figure*}
Considering that the ended projects have lost funding abilities, only running projects have the potential to compete with others~\cite{lin2018modeling}. The competitiveness of a running project is represented by its funding ability in the near feature. So, given the target project $g$, whose funded time is $T_g$. We choose each project $i$ running in the market at time $T_g$, the competitiveness of $i$ can be quantified by Long Short-Term Memory networks (LSTM)~\cite{hochreiter1997long} as:
\begin{align}
\mathbf{h}_{i}^{c} = \mathrm{LSTM}\left(\mathbf{TS}_i\right),\ i \in {\Psi}_g ,
\end{align}
where ${\Psi}_g$ is the set which contains running projects, and we predict the future fund-raising status of each project. Here we utilize
$\mathbf{TS}_i=\left[\ {\xi}_0, {\xi}_1,..., {\xi}_{23}\ \right]$ to represent the hourly time series of project $i$ during the last day before $T_{g}$. In our work, since the investment sequence of project $i$ is
$S_i = \left\{ \langle v_1, t_1 \rangle , \langle v_2, t_2 \rangle , ... , \langle v_{\vert S_i \vert}, t_{\vert S_i \vert} \rangle \right\}$, the hourly amount can be calculated as:
\begin{align}
\begin{split}
{\xi}_k = log_{2}\left(\sum v_j\right), T_{g}-(k+1)*\Delta \leq t_j < T_{g}-k*\Delta\\
\end{split}
\end{align}
where $k = 0,1,...,23$, $\Delta$ represents the time interval of one hour. In our next section, we will use  $\mathbf{h}_{i}^{c}$ as the state of node $i$ in the influence graph. But this approach will bring about a serious problem. LSTM although promotes better results by making more precise state quantifications, once meeting large-scale graphs, such a recurrent structure would lead to a lot of time load. That is because the calculation process of LSTM is slow and cannot be parallelized. To deal with this situation, we propose a precise MLP-based model with prior knowledge of crowdfunding markets.

\subsubsection{MLP-based Quantification with Prior Knowledge}
Multiple Layer Perceptron (MLP), a simple neural network constructed by feedforward multiple fully connected layers (generally 2 or 3 layers) and a nonlinear activation function. MLP is widely used in the large-scale task of machine learning, especially in quite a few applications in the industry. That is because this architecture possesses two great advantages: In theory, it is able to fit any linear or non-linear function; and MLP is essentially a continuous multiplication of several matrices, which is suitable to be parallelized or distributed.
However, using the feedforward network to predict time series is usually unsatisfactory. 
Since it lacks the ability to make use of the order and temporary dependency of the sequence without recurrent architectures and convolution filters. So, the most important information about the trends (or patterns) in time series cannot be learned by MLP. To solve the dilemma, researchers usually incorporate contextual information into the input feature. But in our task, due to many projects that have not been released for a long time, it is not sufficient for the capturing of contextual information.

\begin{figure}[t]	
	\centering
	\begin{subfigure}[t]{.3\columnwidth}
		\centering
		\includegraphics[width=\columnwidth]{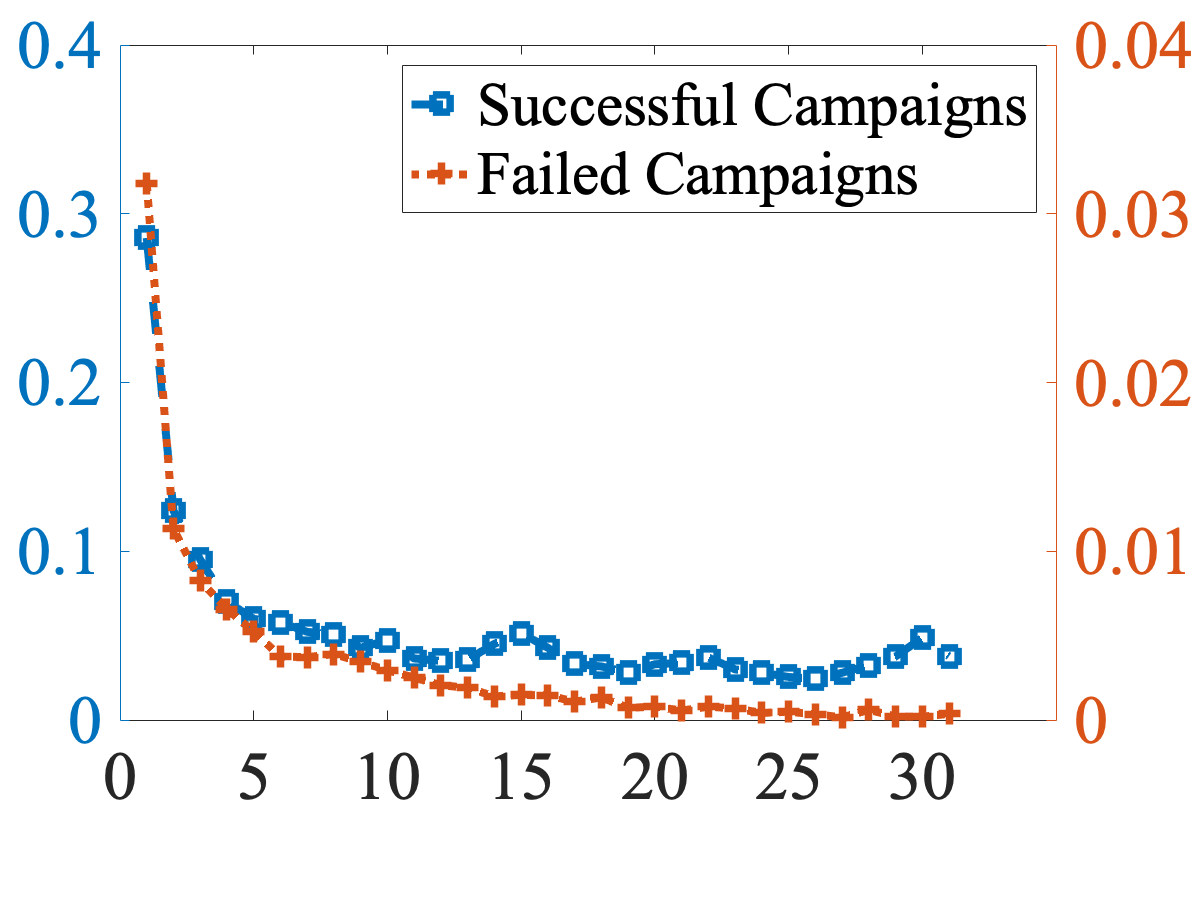}
		\caption{}\label{fig:prior}		
	\end{subfigure}
	\quad
	\begin{subfigure}[t]{.52\columnwidth}
		\centering
		\includegraphics[width=\columnwidth]{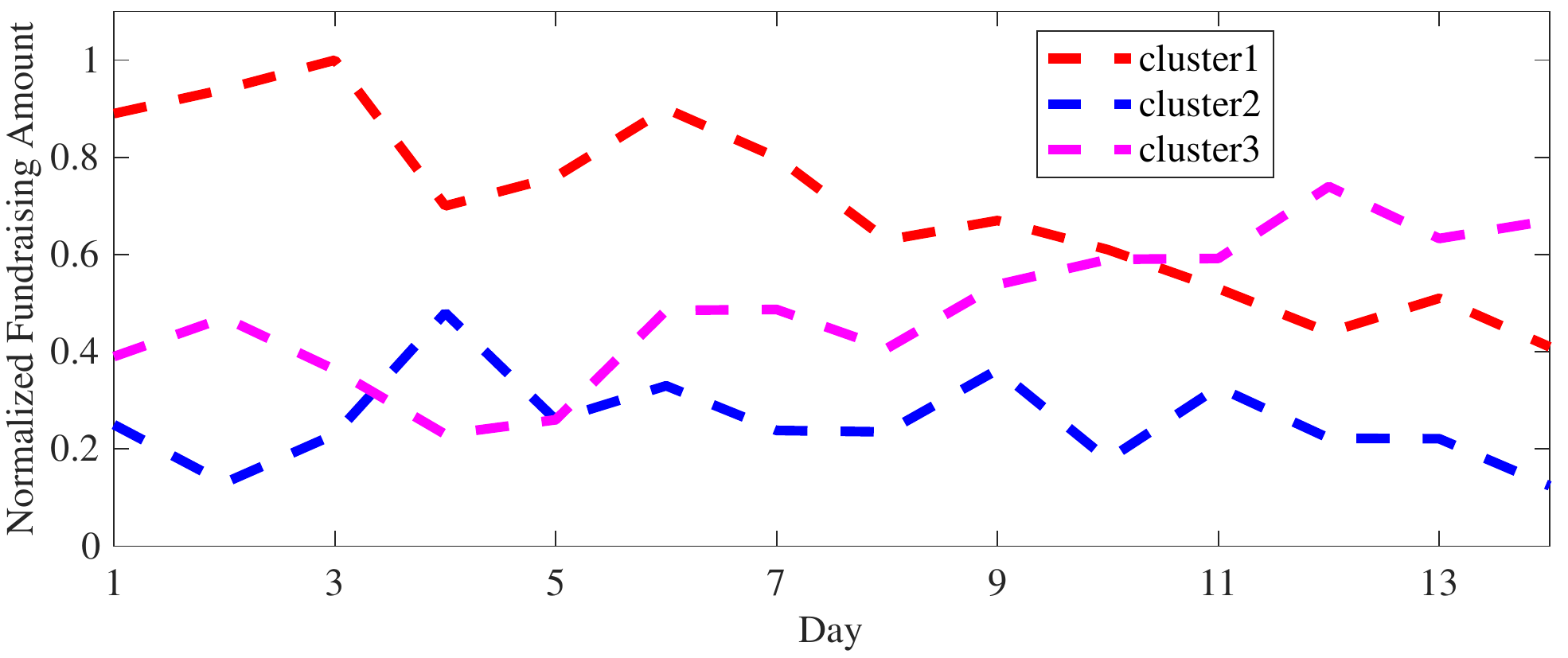}
		\caption{}\label{fig:dynamicprefer}
	\end{subfigure}
	\caption{Data explorations on the dataset. (a). Average daily gain of funding percentage of successful and failed projects. (b). Daily fund-raising amount of each project cluster in one category (Technology).}  
\end{figure}

Fortunately, in the crowdfunding market, there exists an interesting and important phenomenon: Over the overall period, the fund-raising ability in proportion to the fund-raising amount and inversely proportional to the fund-raising days. The herd effect induces investors to incline to choose projects that have been funded more~\citep{shneor2019reward}, and the fund-raising performances of most projects achieve the peak in early stage and then show decreasing trend~\citep{lin2018modeling}. To verify this point, we conduct a data exploration experiment in our Indiegogo dataset. We observe the average daily gain of funding percentage to goals of successful and failed projects. Coinciding with previous researchers, according to Figure~\ref{fig:prior}, we find that curves of average daily gain of successful and failed projects show the attenuation on the overall trend. The fund-raising ability reaches the maximum value at the beginning of the project time and its fluctuation is relatively gentle in the tail. The little fluctuation in the tail would not contribute much to the subsequent process of  competitiveness pressure aggregation, since the rival shows weak competitiveness if it already goes to this stage. It should be noted that we only use the training part, and we assume that the data distribution of the training set and test set tends to be consistent. 

Naturally, we can regard this regular pattern as the priority knowledge of our problem. It shows the varying trend of most projects' fund-raising processes. Such a priority knowledge can be used to adjust the trends learning of our predictors. In particular, we first formulate the impact on trends learning from the priority knowledge with a concise formula:
\begin{align}
\label{prior_define} Tr_i = \frac{\alpha_i / g_i }{log_2 (d_i + 1)}, \ i \in {\Psi}_g,
\end{align}
where $\alpha_i / g_i$ means the achieved process of project $i$, $d_i$ is the funded days. Then, we convert the continuous value to the one-hot vector $\mathbf{V}_p$ to feed into the MLP network. In detail, [0, 1] is divided into several subintervals of equal length, and the value of a sample is mapped into corresponding subinterval. For instance, if the length of one-hot vector is 5,  $Tr_i = 0.1$, we get $\mathbf{V}_{p}^{i} = [1,0,0,0,0] ^\top$. In our work, the dimension of $\mathbf{V}_p$ is set to 6 (best tuning parameter), and there are 3 layers in MLP. With the help of above analysis, we quantify the competitiveness of project $i$ by prior knowledge method as $\mathbf{h}_{i}^{c} = \mathrm{MLP}(\mathrm{concat}(\mathbf{TS}_i, \mathbf{V}_{p}^{i})) $. We will compare these two different competitiveness quantification approaches in the experiment.

\subsubsection{Competitiveness Pressure Aggregation}
Firstly we give the definition of our graph from the view of nodes as:
\begin{myDef}
While creating the competitiveness graph, for every target node $g$ to be predicted, edges $e \in $ $\{ \langle i, g \rangle \mid i \in {\Psi}_g \}$ $($ ${\Psi}_g$ denotes corresponding running set $)$ are used to aggregate rivals' influences. Here the state of each node $i$ except for target nodes is represented by corresponding embedding vector $\mathbf{h}_{i}^{c}$ by the competitiveness quantification module above. Since there is no time series of target node $g$, feeding $\mathbf{x}_g$ into a fully-connected layer to get the embedding vector of $g$. 
\end{myDef}

In a real-world scenario, we should take the huge computational pressure into account while the online platforms face a large number of visits. So, it is not able and necessary to satisfy the demands of users for too precise time point. We expect to divide pending projects into several groups to train and eval. Consequently, referring to the general lifestyle of most humans~\cite{wu2017sequential}, i.e. ``8:00$\sim$12:00'', ``12:00$\sim$14:00'', ``14:00$\sim$17:00'', ``17:00$\sim$20:00'', ``20:00$\sim$24:00'' and ``0:00$\sim$8:00''. Here we define a $\textit{target set}$~$\mathcal{G}$ which contains unpublished projects whose pre-funded time in the same period of one day, and we train these projects in $\mathcal{G}$ simultaneously on one graph. In addition, we regard the observation time point of a target set 
$\mathcal{G}$ as $T_{\mathcal{G}} = min\left\{ T_i \mid i \in \mathcal{G} \right\}$ while we are constructing competitiveness graph. Since it is necessary to promise that the investments after $T_g$ cannot be leaked in advance.
 
However, it is time-consuming to track all time series when meeting a huge running project set $\Psi$, specifically for LSTM. With a deep analysis of online crowdfunding, for a project that just starts less than one day, it would compete for one investor's attention with other projects that are also observed by this investor. Actually, the newly published projects are most likely to be noticed by an investor in the \textit{just-funded} section and corresponding \textit{category} section. Those target projects would be shown in the first few pages of these two sections. Along this way, we design a pruning approach which is represented by one adjacent matrix $\mathbf{A}^G\in{\mathbb R}^{\vert {\mathcal{G}} \vert \times {\vert {\Psi}_g \vert}}$ as:
\begin{align}
\mathbf{A}^{G}_{{\mathcal{M}}_i,{\mathcal{M}}_j} = \left\{
\begin{aligned}
&1,\ \textrm{if}\ T_i - T_j \leq 3\ \textrm{days}\\
&1,\ \textrm{if}\ C_i = C_j\\
&0,\ \textrm{otherwise},
\end{aligned}
\right.
\end{align}
where the original project id is mapped to a unique column index (0 to $\vert {\Psi}_g \vert$) of matrix $\mathbf{A}^G$ by a mapping function $\mathcal{M}$. $\mathbf{A}^{G}_{{\mathcal{M}}_i,{\mathcal{M}}_j}$ indicates whether $j$ connects to $i$, $C_i$ and $C_j$ denote the categories of $i$ and $j$. There are two benefits of using this strategy. Firstly, it would reduce the number of time series need to be processed.  Secondly, the pruning approach filters some noises which do not contribute to the network, and we conduct a corresponding ablation experiment to prove its contribution.

Naturally, among all neighbors, the target project $g$ would be impacted mostly by the one with similar content to $g$ or possesses strong fund-raising ability~\cite{lin2018modeling}. Our method extends Graph Attention Network (GAT)~\cite{velivckovic2017graph} to aggregate neighborhoods' states of $g$ as:
\begin{gather}
e_{gi} = \mathbf{V}^\top \left[\ \mathbf{W}\mathbf{x}_g \parallel \mathbf{W}\mathbf{x}_i\ \right], g \in \mathcal{G},\\
 \begin{split}
 {\alpha}_{gi}  &= {\mathrm{softmax}}\left(e_{gi}\right) = \frac{{\mathrm{exp}}\left({\mathrm{LeakyReLU}}\left(e_{gi}\right)\right)}{\sum_{j \in \mathcal N_g } {\mathrm{exp}}\left({\mathrm{LeakyReLU}}\left(e_{gj}\right)\right)},
 \end{split}
\end{gather}
where $\cdot ^\top$ is the transposition operator, and $\parallel$ denotes concatenating. We extract $\mathcal N_g$ from $\mathbf{A}^{G}_{\mathcal{M}_g:}$, which contains the neighbors of target project $g$.  Moreover, the normalized attention coefficients ${\alpha}_{gi}$ are utilized to calculate a linear combination of the features corresponding to them, to serve as the final output features for node $g$:
\begin{align}
\mathbf{H}_{g}^{c} = \sum_{i \in \mathcal N_g}\left({\alpha}_{gi}\mathbf{W}_{h}\mathbf{h}_{i}^{c}\right), g \in \mathcal{G},
\end{align}
where ${\alpha}_{gi}$ are calculated from static content features. But it is different from GAT, we utilize the product of attention weights ${\alpha}_{gi}$ and predicted financing state $\mathbf{h}_i^{c}$ not $\mathbf{x}_i$. According to this, our model gives consideration to two different influences including fund-raising ability and contents of projects simultaneously.

\subsection{Market Evolution Tracking}
As everyone knows, there is no immutable and frozen market, and the environment of the crowdfunding market is no exception. Indeed, people's investment preferences are changing at every moment. For example, the most popular good in the platform may be the energy-economizing fan in hot summer, but the investors would more interested in novel warming equipment with the coming of winter. 
Particularly we use k-means to cluster Indiegogo's projects belong to one category(Technology) into three clusters based on their contents. We then observe the amount of funds received from the market by each cluster of projects every day. Here we show the first two weeks of Feb. 2020 in Figure~\ref{fig:dynamicprefer}. Obviously, the investment preference of each type of innovation is continuously changing, and the content of cluster3 became more and more popular these days. 

Existing studies~\cite{wu2017sequential,wang2018learning} tends to capture the users' preference evolution in a platform via an LSTM-based model, which processes the sequence that consists of previous items' attraction levels.
However, it is difficult for us to model the dynamic evolution of the crowdfunding market due to its huge scale. Since the online market usually produces hundreds of new projects within a matter of days. Ordinary chain models like RNN or LSTM have no power of processing such a long sequence. To solve this dilemma, we propose a propagation tree-structured GNN with message passing. This framework can not only capture evolution like a chain model, but also deal with large-scale market capacity, and we call it Market Evolution Tracking (MET).

\begin{figure*}[t]
\centering
\includegraphics[width=.98\textwidth]{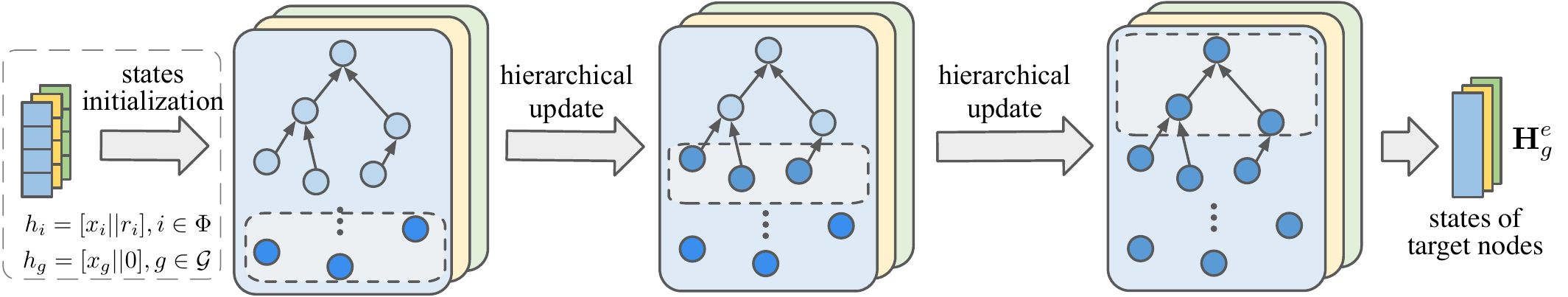} 
\caption{The overview of Market Evolution Tracking (MET) which tracks the past market evolution. $\mathcal{G}$ denotes target set, $\Phi$ denotes observable projects set. To avoid the information redundancy, we propose the novel message passing with hierarchical updating mechanism to transfer the node states from leaves to the target project (root).}
\label{fig:dynamic}
\end{figure*}
\subsubsection{Propagation Tree Constructing}
Actually, from the perspective of situation modeling, the market environment can be regarded as the context of a project. It is worth exploring how the fund-raising process of an individual is affected by the context. For the project $i$ in a market, we should refer to the fund-raising states of other projects in the previous market. We initialize the state of published project $j$ as $\mathbf{h}_{j} = \left[\ \mathbf{x}_{j} \parallel \ {r}_{j}\ \right]$,
\begin{align}
{r}_j = log_2\left(1 + \sum_{t_l < T_j + \mathcal{T}\Delta}{v_l}\right),\ v_l \in S_j,
\end{align}
where the fund-raising stage length $\mathcal{T}$ indicates the time interval, project $j$ attracts the investment amount of $r_j$ in the early stage. We set a constraint that $T_{i} - T_{j} > \mathcal{T}*\Delta$, in this manner $i$ can use message passing to refer to the fund-raising performance of published project $j$ through the connection $\langle j, i \rangle$, here we call $j$ the \textit{observable node} of $i$. For past $t_h$ days, the set of published projects is ${\Phi}_i = \{ j \mid j<N,  \mathcal{T} \ast \Delta  <T_i - T_j< \mathcal{T} \ast t_h \Delta \}$. For the history market learning, the simplest method firstly calculates the funding amount of all nodes in ${\Phi}_i$, and then build all the directed edges from nodes in ${\Phi}_i$ to node $i$.
Unlike the PCM module, we could not directly aggregate all the funding information of nodes in ${\Phi}_i$. This method thinks that the contributions caused by these observable projects are not influenced by their created time. But in fact, there are long-term and short-term effects in general forecasting tasks, which is necessary to assign different reference weights for states with different distances to the pre-startup time point. In our work, we design a message passing based tree-structured GNN to attain the purpose. A toy example can express the idea.
\begin{figure}[t]	
	\centering
	\begin{subfigure}[t]{.3\columnwidth}
		\centering
		\includegraphics[width=\columnwidth]{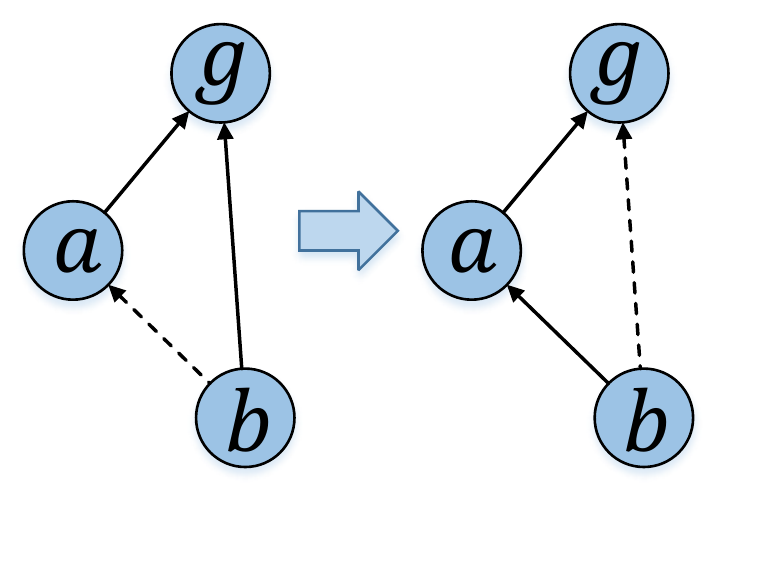}
		\caption{}\label{fig:messa}		
	\end{subfigure}
	\quad
	\begin{subfigure}[t]{.5\columnwidth}
		\centering
		\includegraphics[width=\columnwidth]{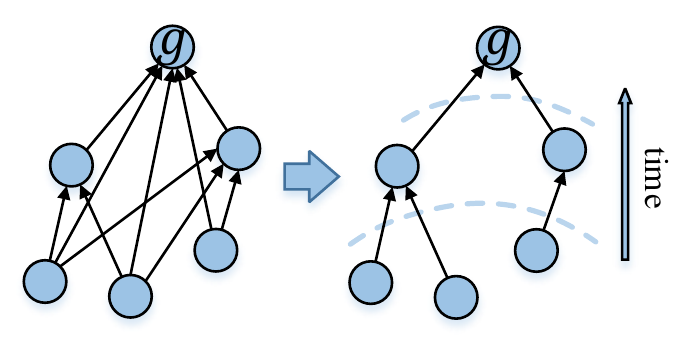}
		\caption{}\label{fig:messb}
	\end{subfigure}
	\caption{The propagation tree structure with message passing. (a). At time $T_g$, node $a$ and $b$ are the observable nodes of target $g$ (pre-startup project). (b). The propagation tree is builded via Algorithm 1.}  \label{fig:message}
\end{figure}

\begin{myExa}
Here we describe our approach which reflects the different levels on time dimension in the market. As shown in Figure~\ref{fig:messa}, the nodes $g$, $a$, $b$ and three edge pairs $\langle a, g \rangle$, $\langle b, g \rangle$, $\langle b, a\rangle$ represent a mini refer market context of target $g$. $a$ is the observable project of $g$, and $b$ is observable for $a$, since per edge lasts more than $\mathcal{T}$ hours. After deleting the edge $\langle b, g \rangle$, we get a tree structure whose root is $g$. Meanwhile, the depths of $a$, $b$ are 1 and 2, which indicate two different time spans to time point $T_g$. It is easy to find that the information transfers from $b$ to $a$ to $g$ via message passing like the step-by-step  propagation of LSTM.
\label{exp:tree}
\end{myExa}

\begin{algorithm}[t]
\caption{Propagation Tree Construction}
{\bf Input:} \\
\hspace*{0.23in}Set of target projects $\mathcal{G}$;\\
\hspace*{0.23in}Set of published observable projects $\Phi$;\\
\hspace*{0.23in}Historical days $t_h$;\\
\hspace*{0.23in}Mapping function ${\mathcal{M}}_i \to \rm{unique\ id}\ in\ \{0,1,...,\vert \Phi \cup \mathcal{G} \vert \},\ i < N $;  \\
{\bf Output:} \\
\hspace*{0.23in}Output adjacency matrix $\mathbf{\Gamma}$
\begin{algorithmic}[1]
\State Set matrix $\mathbf{\Gamma}$ zeros, define infinite matrix $\mathbf{Len}$
\State $U^{0} \gets \emptyset$, $U^{1} \gets \mathcal{G}$
\For{$k=1,...,t_h$}
    \State $B \gets \Phi \cup \mathcal{G} - U^k$
    \For{$i \in B$}
        \For{$j \in U^k$}
            \If{$\mathcal{T} * \Delta  < T_j - T_i < \mathcal{T} * 2\Delta$}
            	 \State $\mathbf{Len}_{j,i} \gets T_j - T_i$
                \If{$k=1$}
                    \State $\mathbf{\Gamma}_{{\mathcal{M}}_j,{\mathcal{M}}_i} \gets 1$ ~~$\rhd$ connect with roots
                \EndIf
            \EndIf
        \EndFor
        \State $ J \gets {\arg\min}_{j\in U^k} (\mathbf{Len}_{j,i})$
        \State $\mathbf{\Gamma}_{{\mathcal{M}}_j,{\mathcal{M}}_i} \gets 1$
        \State $U^{k+1} \gets U^{k} \cup i$
    \EndFor
\EndFor
\State \Return $\Gamma$
\end{algorithmic}
\label{alg:tree}
\end{algorithm}

If we extend the idea of Example~\ref{exp:tree} to a more complex context (Figure~\ref{fig:messb}), there will be some redundant edges that lead to multiple transport paths for a state of sub-node to root. Inspired by the idea of Tree-Lstm~\citep{tai2015improved} in natural language processing, we propose a Propagation Tree Construction algorithm to avoid this dilemma. See \textbf{Algorithm}~\ref{alg:tree} for the detail. Particularly, we train the target set $\mathcal{G}$ simultaneously as the competition module, and each $T_g$ is $\textrm{min}\{T_i| i \in \mathcal{G}\}$. The output of this algorithm is the propagation tree presented in the form of adjacency matrix $\mathbf{\Gamma}$. 
And the depths of nodes represent different time levels. As mentioned before, the market environment can play an important role in representing the context of a target project. For contextual history learning problem~\cite{wu2017sequential}, it is crucial for the long-term trends learning to sample the items of history sequence in a regular interval way. Therefore, while building the edges of node pairs, for each node $i \in \{U^{k}-U^{k-1}\}$, we reserve the shortest edge in $\{\langle j, i \rangle \mid \forall j \in U^{k-1} \}$ to design a needed tree. Along this line, the time intervals between neighbouring layers of this tree are as similar as possible, which leads to each chain represented by a path from one leaf to root that approaches an evenly spaced time series.
\subsubsection{Message Passing with Hierarchical Updating}
Ordinary message passing is not able to support the number of iterations that the tree-structured network requires. We utilize the information propagation method in GG-NNs~\citep{li2015gated} which unrolls the recurrence for fixed steps $t_h$, and uses the backpropagation through time to compute gradients. Firstly, when the time order $t = 1$, all node states are initialized by the concatenation of corresponding static feature and fund-raising amount (the number of target projects are padded with 0) as follow:
\begin{align}
\begin{split}
\mathbf{h}_{i}^{1} &= \left[\ \mathbf{x}_{i} \parallel r_i\ \right],\ i \in \Phi,\\
\mathbf{h}_{g}^{1} &= \left[\ \mathbf{x}_{g} \parallel 0\ \right],\ g \in \mathcal{G}.
\end{split}
\end{align}

Since in the procedure of message passing,  the target nodes and sub-nodes update their states in the same way. For convenience, we use $v$ to denote the id of each node. We define the information aggregation of each node $v$ in each time step as:
\begin{align}
\begin{split}
\mathbf{a}_{v}^{t} = \mathbf{\Gamma}_{\mathcal{M}_v:}^{\mathrm{T}}\left[\ \mathbf{h}_{1}^{(t-1)\mathrm{T}}...\mathbf{h}_{\vert G \cup \Phi \vert}^{(t-1)\mathrm{T}}\ \right] + \mathbf{b}. \\
\end{split}
\end{align}

\begin{algorithm}[t]
\caption{Hierarchical Update Algorithm}
{\bf Input:} \\
\hspace*{0.23in}Set of target projects $\mathcal{G}$,
historical days $t_h$;\\
\hspace*{0.23in}Mapping function ${\mathcal{M}}_i \to \rm{unique\ id}\ in\ \{0,1,...,\vert \Phi \cup \mathcal{G} \vert \},\ i < N $;  \\
\hspace*{0.23in}Set $U^0,U^1, U^2,..., U^{t_h}$ in Algorithm 1,
adjacency matrix $\mathbf{\Gamma}$. \\
{\bf Output:} \\
\hspace*{0.23in}Final state $\mathbf{H}_{g}^{e}$ of node $g \in \mathcal{G}$
\begin{algorithmic}[1]
\State Initialize the node state $\mathbf{h}_{i}^{1} = \left[\ \mathbf{x}_{i} \parallel r_i\ \right],\ i \in \Phi; \mathbf{h}_{g}^{1}= \left[\ \mathbf{x}_{g} \parallel 0\ \right],\ g \in \mathcal{G}$.
\For{$t=2,...,t_h+1$}
    \State Get the nodes whose depths are $d=t_h-t+1$, $U=U^{d+1} - U^{d}$
    \For{each node $v$ in $U$}
    	\State $\mathbf{a}_{v}^{t} = \mathbf{\Gamma}_{\mathcal{M}_v:}^{\mathrm{T}}\left[\ \mathbf{h}_{1}^{(t-1)\mathrm{T}}...\mathbf{h}_{\vert G \cup \Phi \vert}^{(t-1)\mathrm{T}}\ \right] + \mathbf{b}$
	\State Update $\mathbf{h}_{v}^{t}$ as Eq.12
    \EndFor
    \For{each node $v$ not in $U$}
    	\State $\mathbf{h}_{v}^{t}=\mathbf{h}_{v}^{t-1}$  ~$\rhd$ Nodes whose depths are not $d$ needn't update
    \EndFor
\EndFor
\State \Return Final state $\mathbf{H}_{g}^{e} = \mathbf{h}_{g}^{t_h},\ g \in \mathcal{G}$
\end{algorithmic}
\label{alg:update}
\end{algorithm}
We implement the gated recurrent unit to guarantee the long-step propagation of information, this unit will learn to update the newly necessary information within the training process. In detail, the gated module for each node at $t$ as follows:
\begin{align}
\begin{split}
\mathbf{z}_{v}^{t} &= \sigma\left(\mathbf{W}_{z}\mathbf{a}_{v}^{t} + \mathbf{U}_{z}\mathbf{h}_{v}^{t-1}\right), \\
\mathbf{r}_{v}^{t} &= \sigma\left(\mathbf{W}_{r}\mathbf{a}_{v}^{t} + \mathbf{U}_{r}\mathbf{h}_{v}^{t-1}\right), \\
\mathbf{h}_{v}^{t'} &= {\mathrm{tanh}}\left(\mathbf{W}\mathbf{a}_{v}^{t} + \mathbf{U}\left(\mathbf{r}_{v}^{t} \odot \mathbf{h}_{v}^{t-1}\right)\right),\\
\mathbf{h}_{v}^{t} &= \left(1 - \mathbf{z}_{v}^{t}\right) \odot \mathbf{h}_{v}^{t-1} + \mathbf{z}_{v}^{t} \odot \mathbf{h}_{v}^{t'}.
\end{split}
\end{align}

After $t=t_h$, we get the final state of target $g$ which has learned the evolution pattern of propagating from entire past market,
\begin{align}
\begin{split}
\mathbf{H}_{g}^{e} &= \mathbf{h}_{g}^{t_h},\ g \in \mathcal{G}.\\
\end{split}
\end{align}

However, following the regular rule of message passing, if we renew all nodes in the tree at each timestep, there is a non-ignorable problem that the information of some nodes is aggregated to root several times.  For instance, as shown in Figure~\ref{fig:messa}, the state of node $a$ would be transmitted to root $g$ two times after timestep $t=2$. This situation leads to two disadvantages: $(1)$ It still obeys our aim of creating the propagation tree which divides the historic nodes into numerous evenly spaced time series. We expect step-by-step time series to denote different time levels of nodes, but the root receives the same node's feature several times, which will interfere with the weights of different time levels; $(2)$ A large number of nodes do not need to update their states at every timestep, so it would affect the training efficiency of our model on a large-scale network.

We propose a hierarchical update algorithm to remedy this flaw. In~\textbf{Algorithm}~\ref{alg:update}, we adopt the partial update way which only processes the nodes with a certain depth at time $t, t \in \{1,2,...,t_h\}$. In other words, we transfer the node states of the propagation tree from leaves to root in turn and only update the nodes with a certain depth at each iteration. In this way, it is easy to prevent the transporting of redundant information. To sum it up, comparing with the chain-like sequential models that would degenerate over a long distance, the propagation tree structure with much fewer layers is a better way of modeling the influence propagation of the entire market. Therefore, our propagation tree algorithm is more suitable for modeling the crowdfunding market evolution.

\subsection{Joint Optimization Strategy}
The loss function is designed based on the combination of the output state of PCM and MET module. Firstly, we feed the above two output states into a fully connected layer to get the predicted value as:
\begin{align}
\begin{split}
y_{g}^{'} = {\mathrm{ReLU}}\left(\mathbf{W}_{f}\left(\mathbf{H}_{g}^{c} + \mathbf{H}_{g}^{e}\right) + b_f\right),\ g \in \mathcal{G},
\end{split}
\end{align}
where $\mathbf{W}_{f}$ is the parameter of FC layer, $b_f$ indicates corresponding bias, and ReLU~\citep{nair2010rectified} keeps the predicted value $y_{g}^{'} \geq 0$. Then, 
we define the loss function as:
\begin{align}
\begin{split}
\mathit{Loss_p} = \frac{1}{\vert \mathcal{G} \vert} \sum_{g \in \mathcal{G}} \vert y_g - y_{g}^{'} \vert.\label{eq:lossp}
\end{split}
\end{align}

Naturally, in the competitiveness quantification, the more accurate fitting of rivals' states leads to the better states aggregation. So the MAE loss $\mathit{Loss}_l$ of time series predicting in the competitiveness quantification can be calculated in the same way as Equation (\ref{eq:lossp}), that is $\mathit{Loss_l} = \frac{1}{\vert \Psi \vert} \sum_{i \in \Psi} \vert y_i - y_{i}^{'} \vert$, where $y_{i}^{'} = {\mathrm{ReLU}}\left(\mathbf{W}_{f}^{'}\mathbf{h}_{i}^{c} + b_{f}^{'}\right)$, $y_i$ denotes the ground truth (the $log_2$ value of fund-raising amount in next day) of project $i$ in the running project set $\Psi$, $\mathbf{W}_{f}^{'}$, $b_{f}^{'}$ are the training parameter matrix and bias. We then adopt trade-off to assign the weights between $\mathit{Loss}_l$ and $\mathit{Loss}_p$, and jointly train these two losses. The trade-off hyperparameter is denoted by $\eta$, a float value between 0 and 1. Here the objective function is:
\begin{align}
\begin{split}
\mathit{f}(\Theta) = \min_{\Theta} \left(\eta \mathit{Loss}_p + \left(1-\eta\right) \mathit{Loss}_l\right).
\end{split}
\end{align}

Finally, during the training process, we utilize Stochastic Gradient Decent(SGD)~\cite{dekel2012optimal} to update the parameters $\Theta$ of our model and apply exponential decay to the initial learning rate 0.02.

\section{Experiments}
In this section, we provide all information about our experiments on the fund-raising estimation task. Firstly, we describe the Indiegogo dataset in some detail and express the experimental setup. Then, the introduction of baselines and the results of comparison experiments are given. Finally, we report several verified experiments for major modules in our model.
\subsection{Dataset Information}
Our experiments are conducted on Indiegogo dataset~\cite{liu2017enhancing} which consists of $N = 14,143$ projects and $\vert S \vert = 1,862,097$ investing records.
The projects' heterogeneous information is shown in Table~\ref{table:features} mentioned in preliminaries. Although there is only a public crowdfunding dataset that includes dynamic investment records, we cut three kinds of sub-datasets consist of 7K, 10K, and 14K projects respectively, so as to test the generalization ability of GME. Here each sub-dataset obtains the same features as the whole dataset except for the number of instances. In detail, we randomly choose the start point and the corresponding end point on the timeline to intercept different scale sub-datasets, e.g., start point 1, end point 7000, we catch all the projects and their features from the earliest published project to the 7000th in chronological order. With the split ratio of 5:1, we divide each dataset into a training set and test set in chronological order. Note that, to ensure that the deviation of experimental results is as small as possible, for every certain scale, we repeat the above process 10 times for training and calculate average results. 
Moreover, the historical market data of each target set is captured by a corresponding sliding window along with the time order.
It is normal to use the moving partition strategy with the sliding step( actually, it is the segment period mentioned in section 3.2.3), and the lengths $t_h = \{1,2,...,7\}$ of time windows are adopted in our experiments.

\subsection{Experimental Setup}
\subsubsection{Evaluation Metrics.}
As the same as other regression and prediction models, we use two standard evaluation metrics Mean Absolute Error (MAE) and Root Mean Squared Error (RMSE)~\cite{wang2018learning} to evaluate the proposed model and other baselines. Specifically, $\mathcal{G}$ denotes a target projects set of one time window, and the actual target value $y_i$ and predicted value $y_{i}^{'}$ of project $i \in \mathcal{G}$ are described in preliminaries. The evaluation protocols are defined as follow:
\begin{align}
\mathit{MAE} = \frac{1}{\vert \mathcal{G} \vert} \sum_{i=0}^{\vert \mathcal{G} \vert} {\vert y_i - y_{i}^{'} \vert} , \ 
\mathit{RMSE} = \sqrt{\frac{1}{\vert \mathcal{G} \vert} \sum_{i=0}^{\vert \mathcal{G} \vert}\left(y_i - y_{i}^{'}\right)^{2}} .
\end{align}

\subsubsection{Baselines.}
Since \textit{Fund-raising Performance Estimation} is an open issue, most existing crowdfunding models are inapplicable to the task. 
Therefore, we choose many well-designed and powerful methods from other related tasks (their problem definitions are similar to ours or in the same research field). Some of them are fine-tuned to transfer to adapt to our problem scenario.

\begin{itemize}
\item $\textbf{LR}$ (\textit{Linear Regression}) establishes the relationship between the target variable and the input variable by fitting a straight regression line. It processes the static feature $x_g$.
\item $\textbf{RFR}$ (\textit{Random Forest Regression}) is a supervised learning algorithm which can be used to ensemble meta-learning method~\cite{liaw2002classification}. The random forest algorithm usually gets better results than a single decision tree. The reason is it combines multiple decision trees, and the method can decrease the variance. In our model, we use the MSE criterion to grow the individual decision trees. Since the random forest algorithm requires not much parameter tuning, so we optimize it by setting the max depth of RFR to 4 finally and process static features for prediction.
\item $\textbf{MLP}$ (\textit{Multiple Layer Perceptron}) is a classical feedforward neural network trained with backpropagation \cite{zhang1998comparison}. Given a target $g$, we firstly concatenate the static feature $x_g$, the average pooling embedding of running projects' competitiveness at $T_g$ (\textbf{MLP-C}) and the average pooling embedding of the initial states of nodes in $tree_g$ (\textbf{MLP-H}) together. Next, these concatenated vectors would input into an MLP which contains three fully connected layers ([150, 50, 1]) with the ReLU activation function for predictions.
\item $\textbf{LSTM}$ (\textit{Long Short-Term Memory networks}) is a special extension of recurrent neural network architecture that can learn long-term dependencies and restrain vanishing gradient~\cite{hochreiter1997long}. This architecture is a general sequence modeling tool. We implement LSTM to capture the short-term history of the crowdfunding market. In the compared experiments, the time step of LSTM is set to 15, the dimension of the hidden state is 50.
\item $\textbf{CLSTM}$ (\textit{Contextual LSTM}) regards the contextual feature as part of the input of LSTM, so it would consider the context within the message propagation \cite{ghosh2016contextual}. We design the contextual feature via creating the one-hot encode vector of length 24 for hourly temporal variables.
\item $\textbf{DTCN}$ (\textit{Deep Temporal Context Network}) is a deep learning model which proposed for predicting sequential image popularity \cite{wu2017sequential}. To some degree, the background of this task is a little similar to the prediction problem we propose. This model reveal individual preference and public attention from evolutionary social systems. It incorporates both temporal context and temporal attention into the framework. Here the dimension of temporal context by 24 and the parameter of the chain module as same as the LSTM.
\item $\textbf{GCN}$ (\textit{Graph Convolutional Network}) is a simplified spectral method of node aggregating, we define the neighbors of node $g$ as running projects' competitiveness at $T_g$. The dimension of hidden state is 50.
\item $\textbf{GAT}$ (\textit{Graph Attention Network}) weights a nodes’ neighbors by attention mechanism, in the same way, the neighbors of node $g$ as running projects' competitiveness at $T_g$. The dimension of hidden state is 50.
\item $\textbf{DMC}$ (\textit{Dynamic Market Competition}) is a probabilistic generative model which can distinguish the different competitiveness of projects in crowdfunding markets \cite{lin2018modeling}. The model thinks that the projects in the competition spaces characterized by popular contents would attract more investments than those non-mainstream projects.  We slightly fine-tune the method to make it applicable to our task and make the number of competition spaces 10.
\end{itemize}
Specifically, we conduct an ablation experiment to verify the importance and necessity of the PCM and MET components.
\begin{itemize}
\item $\textbf{GME-C}$ denotes part of the whole model that GME removes the market evolution module. We compare it with other methods for judging the necessity of PCM.
\item $\textbf{GME-H}$ denotes part of the whole model that GME remove the projects competition module. We compare it with other methods for judging the necessity of MET.
\end{itemize}

\subsubsection{Implementation Details.}

\linespread{1.36}
\begin{table}
\tiny
\caption{Comparisons on the datasets.}
\centering
\begin{subtable}{1.\textwidth}
\caption{Fund-raising performances of first $\mathcal{T}=24$ hours.}
\centering
\resizebox{0.95\columnwidth}{!}{
\begin{tabular}{c|cc|cc|cc}
\hline
\multirow{2}{*}{Models} & \multicolumn{2}{c|}{Indiegogo-7K} & \multicolumn{2}{c|}{Indiegogo-10K} & \multicolumn{2}{c}{Indiegogo-14K}\\

\cline{2-7} & MAE & RMSE & MAE & RMSE & MAE & RMSE\\
\hline
LR & 0.2214 & 0.334 & 0.2242 & 0.3331  & 0.223 & 0.3364\\
RFR & 0.2175 & 0.3343 & 0.2224 & 0.3293  & 0.2251 & 0.3376\\
MLP & 0.2201 & 0.3377 &0.2209 & 0.3351  & 0.2193 & 0.3305\\
MLP-H & 0.2235 & 0.3398 &0.2227 & 0.3336 & 0.2225 & 0.3343\\
MLP-C & 0.2273 & 0.3381 &0.2215 & 0.3348 & 0.2283 & 0.3378\\
LSTM & 0.2134 & 0.3316 &0.2238 & 0.3324 & 0.2117 & 0.3240\\
CLSTM & 0.2112 & 0.3279 &0.2128 & 0.3266 & 0.2146 & 0.3293\\
DTCN & 0.2024 & 0.3165 &0.1978 &0.3018  & 0.2044 & 0.3142\\
GCN  & 0.2006 & 0.3135 &0.1986 & 0.3077 & 0.2075 & 0.3139\\
GAT & 0.1918 & 0.3062 &0.2003 & 0.3110 & 0.2070 & 0.3133 \\
DMC & 0.1970 & 0.3100 &0.1956 & 0.3117 & 0.1933 & 0.2938\\
\hline
GME-C & 0.1941 & 0.3047 &0.1974 & 0.3141 & 0.2053 & 0.3078\\
GME-H & 0.1883 & 0.3066 &0.1911 & 0.3007 & 0.1895 & 0.2910\\
\textbf{GME} & \textbf{0.1770} & \textbf{0.2952} &\textbf{0.1797} & \textbf{0.2881} & \textbf{0.1820} & \textbf{0.2863}\\
\hline
\end{tabular}
}

\end{subtable}

\begin{subtable}{1.\textwidth}
\caption{Fund-raising performances of first $\mathcal{T}=48$ hours.}
\centering
\resizebox{0.95\columnwidth}{!}{
\begin{tabular}{c|cc|cc|cc}
\hline
\multirow{2}{*}{Models} & \multicolumn{2}{c|}{Indiegogo-7K} & \multicolumn{2}{c|}{Indiegogo-10K} & \multicolumn{2}{c}{Indiegogo-14K}\\

\cline{2-7} & MAE & RMSE & MAE & RMSE & MAE & RMSE\\
\hline
LR & 0.2278 & 0.3472 & 0.2347 & 0.345  & 0.2404 & 0.3569\\
RFR & 0.2309	&0.3499	&0.2322	&0.3416	&0.2388	&0.356\\
MLP & 0.2288	&0.3463	&0.2334	&0.3497&	0.2336	&0.3495\\
MLP-H & 0.2333	&0.3493&	0.2353&	0.3485&	0.2315&	0.3437\\
MLP-C & 0.2342&	0.3458	&0.2349	&0.3518	&0.2411	&0.3504\\
LSTM & 0.2211	&0.3417&	0.2334&	0.3427&	0.2224&	0.3385\\
CLSTM &0.2254	&0.3483	&0.2193&	0.3344	&0.2292	&0.3494\\
DTCN & 0.2104	& 0.3267	&0.2071	&0.3116	&0.212&	0.3215\\
GCN  &0.2109	&0.3275&	0.209&	0.3215&	0.2218&	0.3309\\
GAT & 0.1996	&0.3156&	0.2089&	0.3225&	0.2104&	0.3187 \\
DMC & 0.2071	&0.3224	&0.209	&0.3248	&0.2024&	0.3052\\
\hline
GME-C & 0.2025	&0.3172&	0.2041&	0.3203&	0.2121&	0.313\\
GME-H & 0.1906	&\textbf{0.305}&	0.2027&	0.316&	0.202&	0.3068\\

\textbf{GME} & \textbf{0.1883} & 0.3077 &\textbf{0.1918} & \textbf{0.3034} & \textbf{0.1929} & \textbf{0.2953}\\
\hline

\end{tabular}
}
\end{subtable}
\label{table:compar}
\end{table}

For text feature, it should be segmented into individual words firstly, and all words present as the lowercase type. Then we remove all punctuations and every word which appears less than 5 times. These text features can be transformed into 50-dimensional vectors via the doc2vec algorithm. When dealing with static numerical features, we implement the discrete data type of one-hot encoder to represent them. Specifically, the dimensions of declared funding duration and declared pledged goal are 4 and 16, respectively. For our model and baselines, the size of the hidden state in LSTM and the state of nodes in graphs are both 50, so the FC layers are a  $50 \times 1$ parameter matrix. Besides, we set the trade-off weight $\eta$ as 0.7 and keep the dropout after the gated module as 0.9 empirically.

\subsection{Experimental Results}
\subsubsection{Analysis of Prediction Performances.}
We show the comparison results of all methods in Table~\ref{table:compar} and they reflect the differences that the models perform on different scale datasets and two different spans of the early stage. As mentioned before, we use the fund-raising performance within the first $\mathcal{T} = 24, 48$ hours because our estimation of fund-raising performance is before the project sets up. This setting aims at eliminating influences of the project changes after setting up.

Overview, our proposed model GME achieves the best scores among three data size settings. Apart from GME-based models (GME-H and GME-C), when we estimate the fund-raising performance within the first 24 (48) hours, GME outperforms other compared approaches with averages of 16.1\%, 15.3\%, and 14.9\% (14.5\%, 13.9\%, and 14.3\%) relative MAE improvements on Indiegogo-7K, 10K and 14K respectively. To go deep into the components of GME, GME-H works better than the  sequential models includes LSTM, CLSTM, and DTCN both on MAE and RMSE. That means the proposed propagation tree is more applicable to learn the market environment evolution although this structure is not complex.  With the same node states, GME-C surpasses the performances of GCN and GAT, which illustrates the superiority of our design. Moreover, from the point of view of the ablation experiment, there is an obvious appearance that the complete model GME surpasses the individual GME-H and GME-C by a distinct margin. It proves that the combination of competition learning and market evolution learning is key to improve prediction performance. In conclusion, our experiments illustrate the effectiveness of modeling the crowdfunding market.
\begin{figure}[t]	
	\centering	
	\begin{subfigure}[t]{0.42\columnwidth}
		\centering
		\includegraphics[width=\columnwidth]{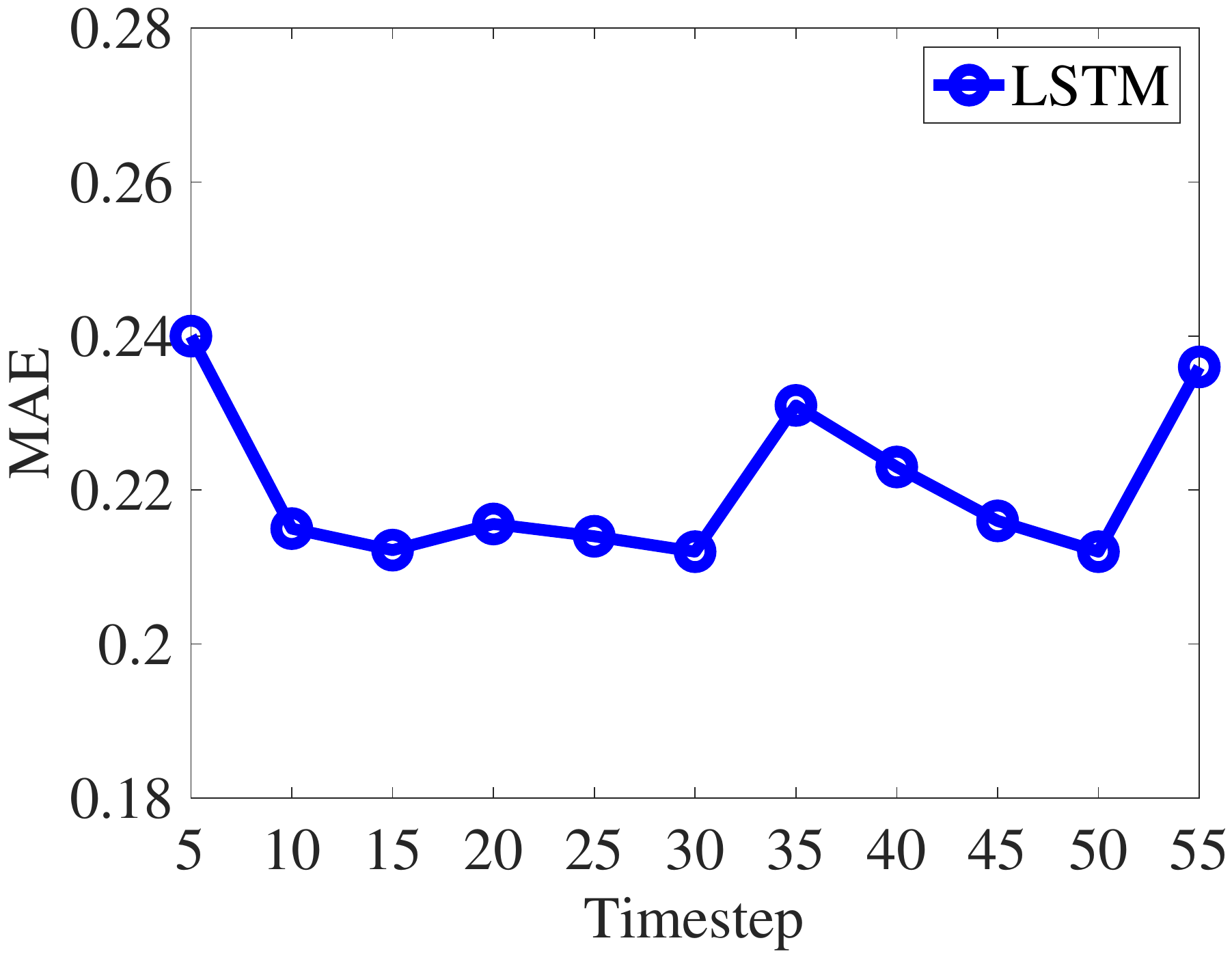}
		\caption{Timestep influence LSTM.}\label{fig:lensb}
	\end{subfigure}
	\quad
	\begin{subfigure}[t]{0.42\columnwidth}
		\centering
		\includegraphics[width=\columnwidth]{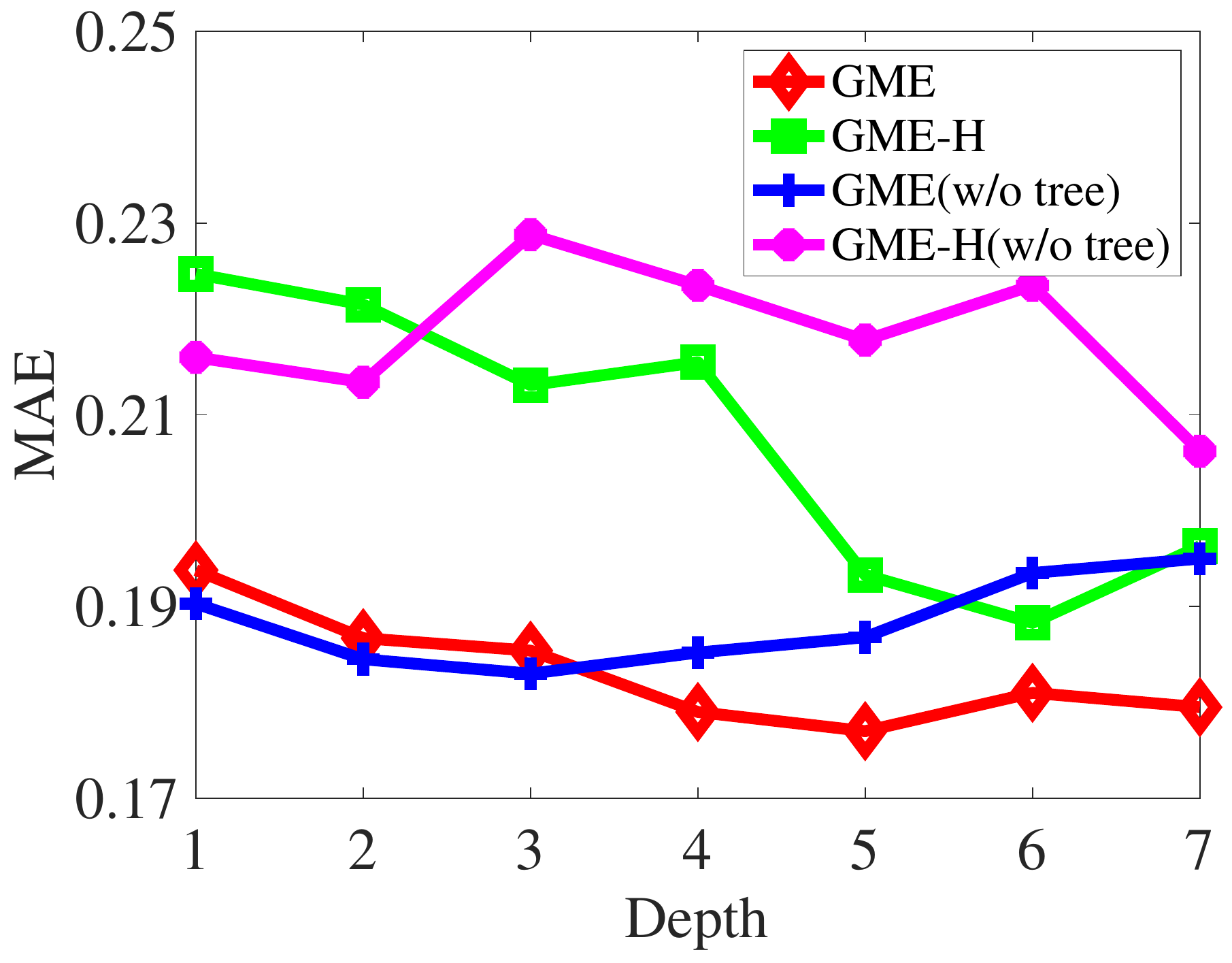}
		\caption{MAE changes with depth on 7K.}\label{fig:lensa}		
	\end{subfigure}
	\begin{subfigure}[t]{0.42\columnwidth}
		\centering
		\includegraphics[width=\columnwidth]{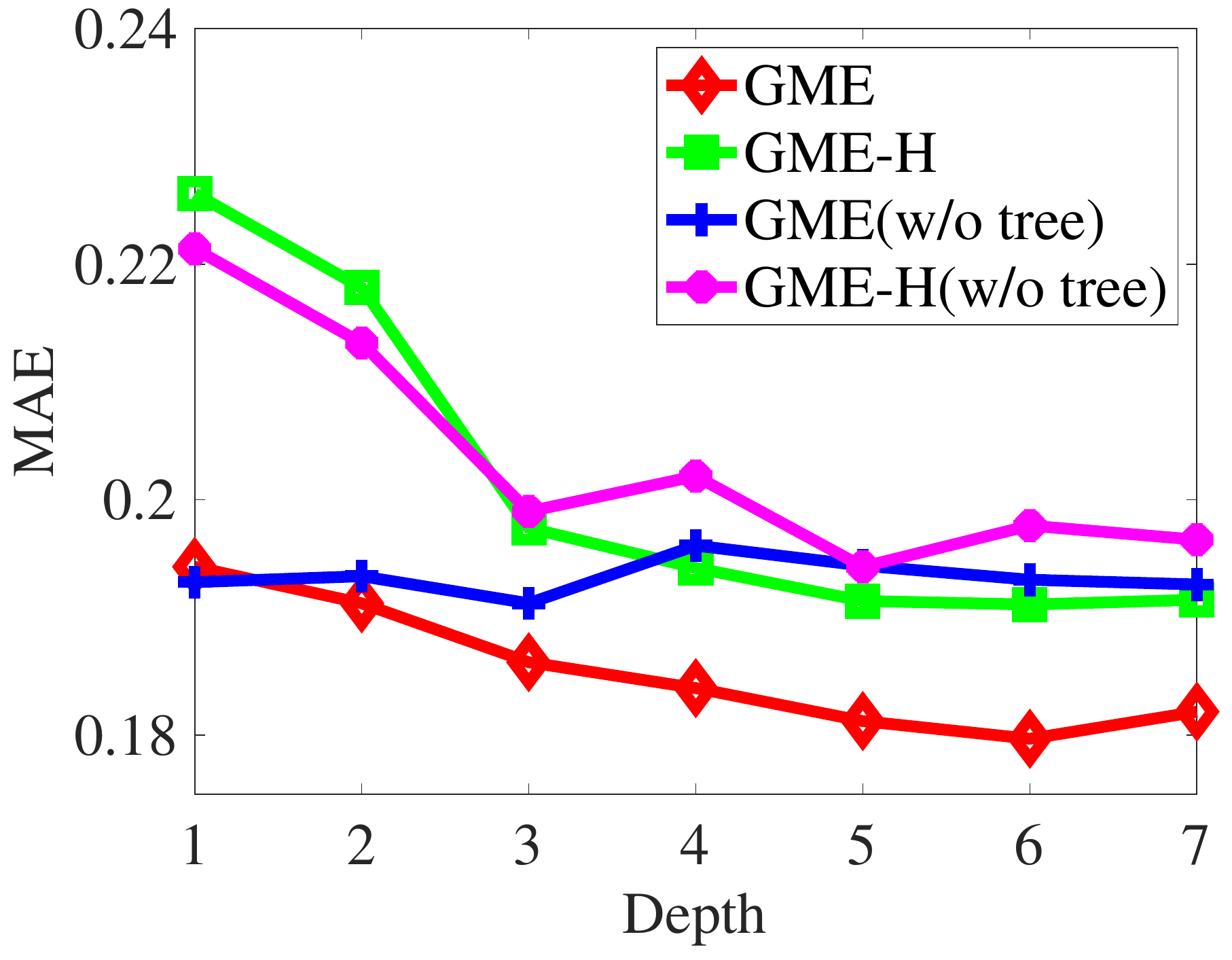}
		\caption{MAE changes with depth on 10K.}\label{fig:lensb}
	\end{subfigure}
	\quad
	\begin{subfigure}[t]{0.42\columnwidth}
		\centering
		\includegraphics[width=\columnwidth]{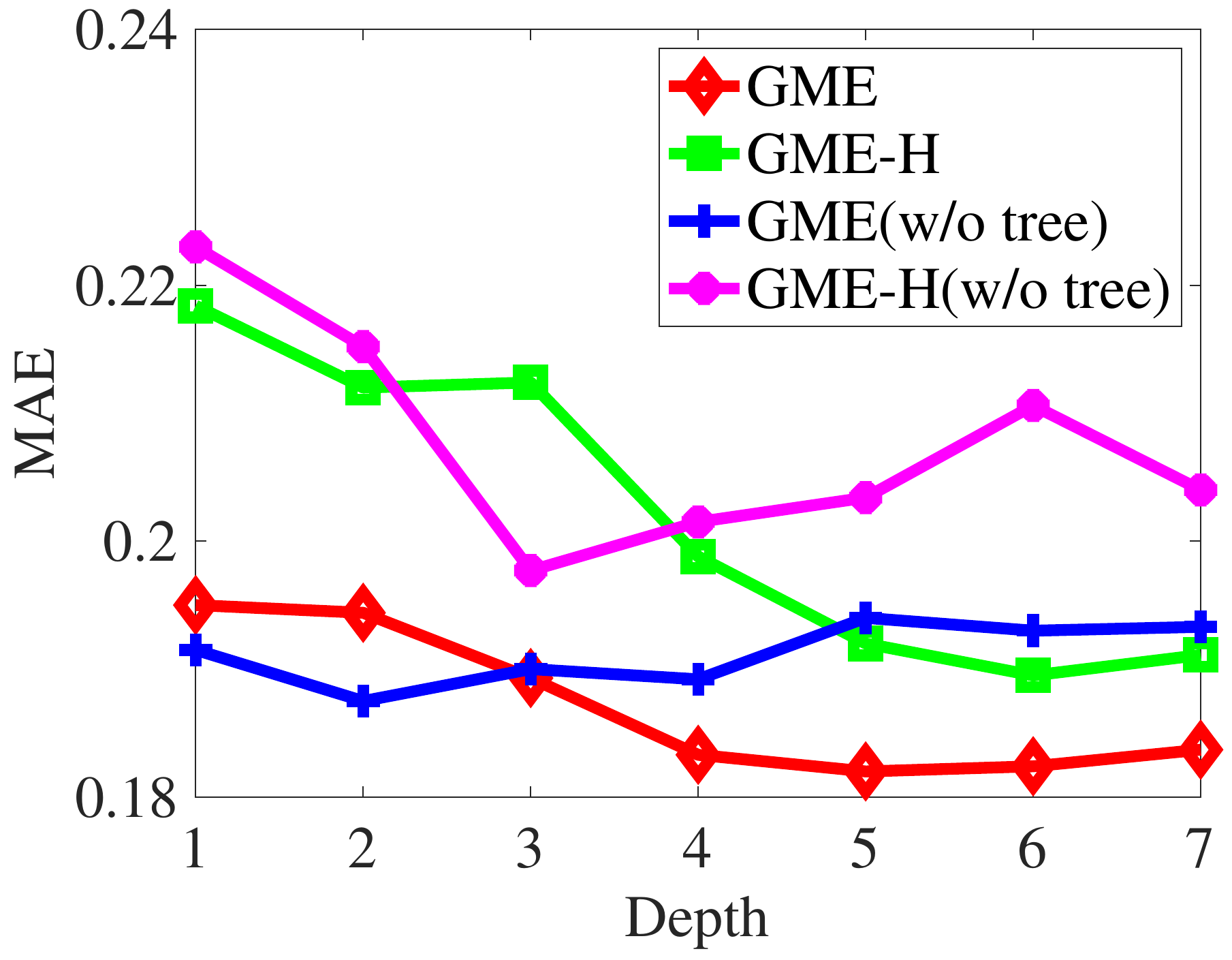}
		\caption{MAE changes with depth on 14K.}\label{fig:lensa}		
	\end{subfigure}
	\caption{Influence of scale of market evolution.}\label{fig:lens}
\end{figure}
\subsubsection{Exploration on the Scale of Market Evolution.}
In our work, the propagation tree is designed for modeling the large-scale characteristic of market evolution. We visualize the influence of this cause in Figure~\ref{fig:lens}. In detail, $\mathcal{T}=24$ in this experiment, and the depth indicates the number of observable history days before the pre-startup time of the target project. We define “w/o tree” to denote the architecture without the tree structure, which aims for explaining the reason why we create the propagation tree rather than directly use GCN to aggregate all nodes in the past market. Analyzing the charts carefully, a noticeable one is that the GME model expresses a stable descent in the overall trend amongst three datasets, and it also drops down to the best point value. Besides, although GME-H causes vibrations on the minimum dataset, its pattern gradually becomes smooth with the increment of data size. We notice that GME and GME-H show slight upward trends at the tail of their curves. 
And in the experiments, it is beneficial for GME to set the depth as 5 or 6, since it performs best with this setting. This is a normal phenomenon that the longer history record may bring about more noise, so it reveals that lengthening history length unlimited would not always lead to the improvement of performance.
On the other hand, GME-H (w/o tree) and GME (w/o tree) cannot work better with the larger depth (number of past days). For instance, we find GME (w/o tree) usually achieves the best score when the depth equals 2 or 3, and the curves of GME-H (w/o tree) are not even able to present an obvious tendency. The gaps between the tree-based models and single time level models indicate the advantage of the propagation tree structure.

\subsubsection{Influence of Pruning Approaches.}
\begin{figure}[t]	
	\centering	
	\begin{subfigure}[t]{0.32\columnwidth}
		\centering
		\includegraphics[width=\columnwidth]{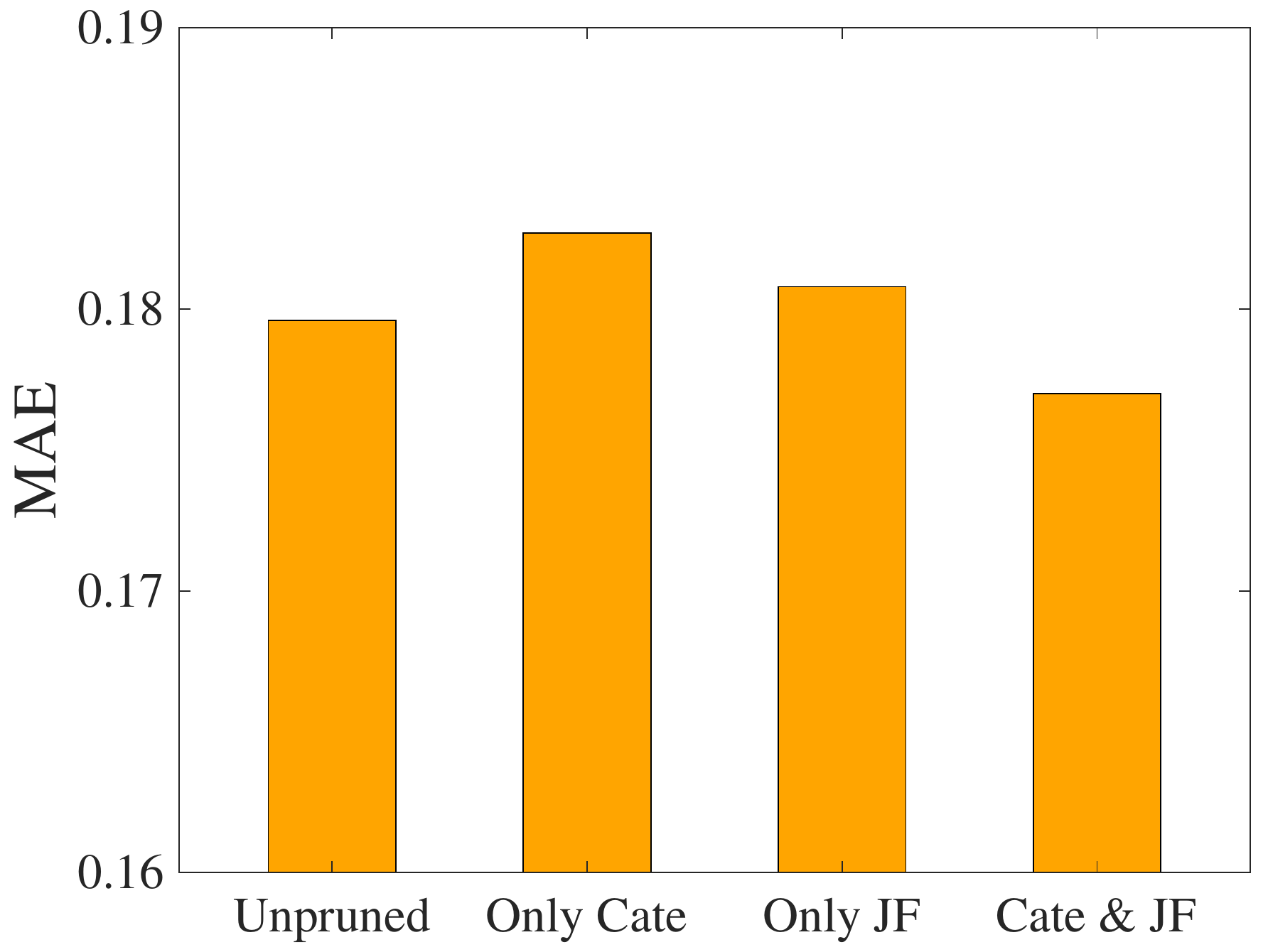}
		\caption{Indiegogo-7K.}\label{fig:lensb}
	\end{subfigure}
	\begin{subfigure}[t]{0.32\columnwidth}
		\centering
		\includegraphics[width=\columnwidth]{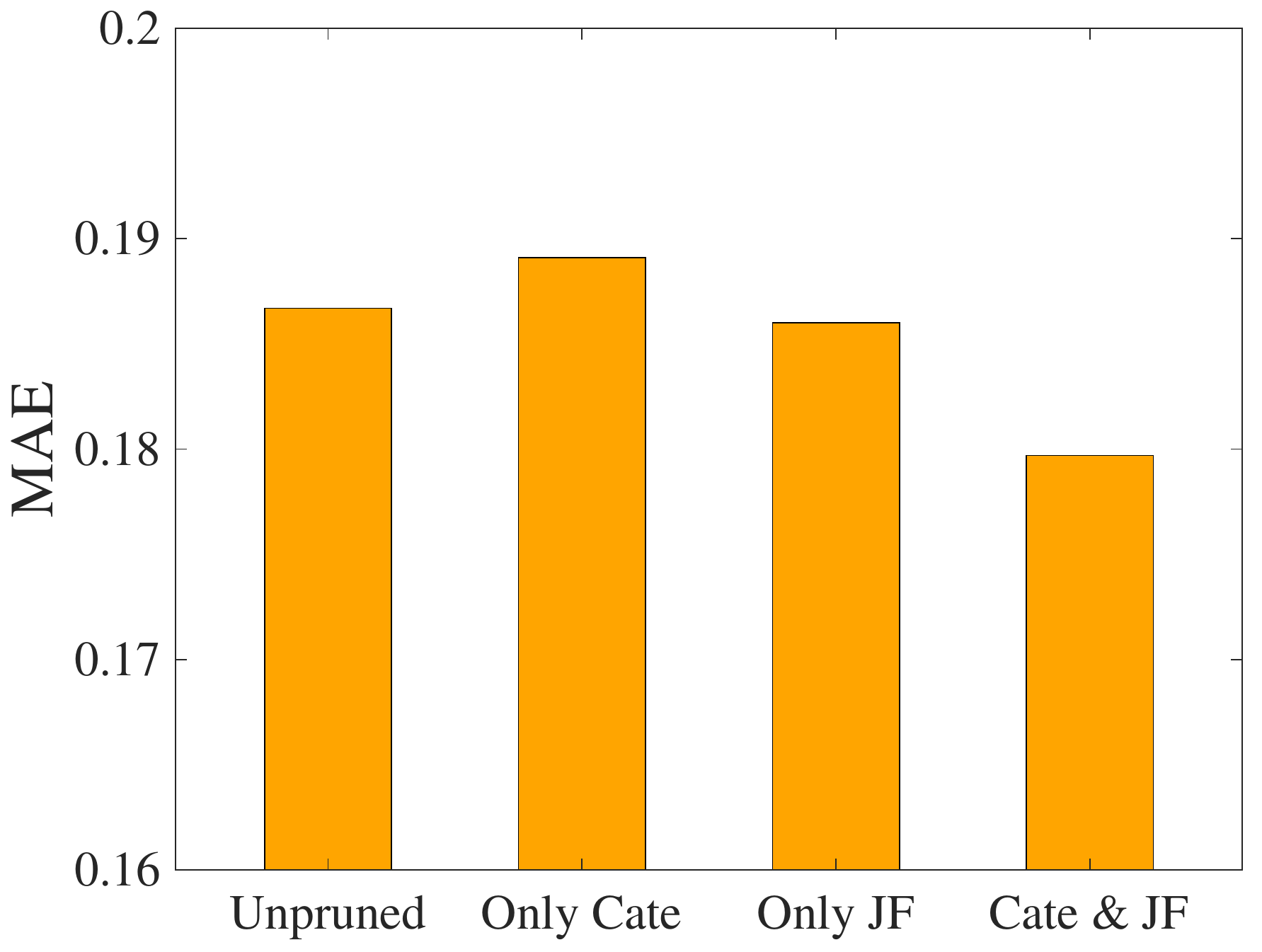}
		\caption{Indiegogo-10K.}\label{fig:lensa}		
	\end{subfigure}
	\begin{subfigure}[t]{0.32\columnwidth}
		\centering
		\includegraphics[width=\columnwidth]{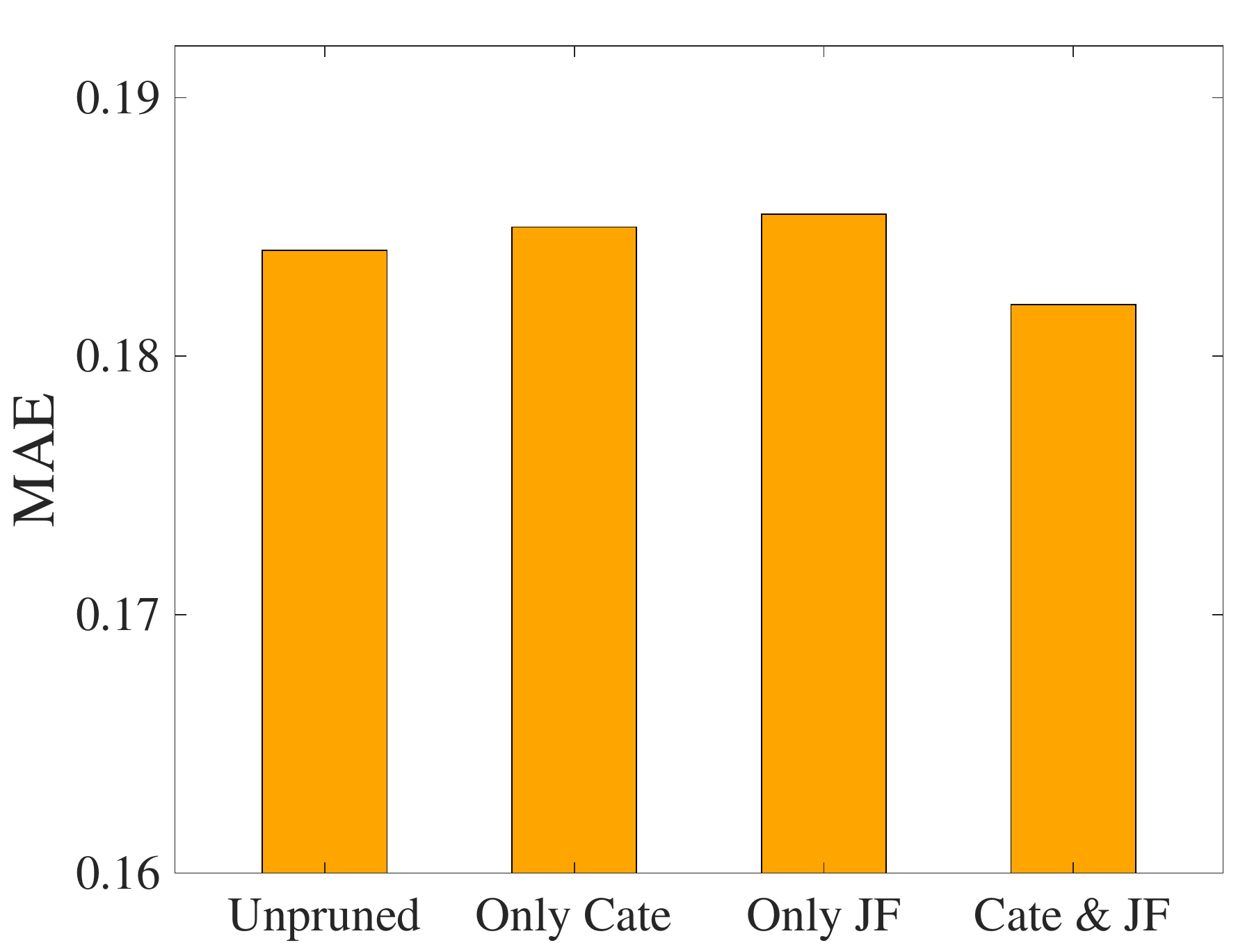}
		\caption{Indiegogo-14K.}\label{fig:lensb}
	\end{subfigure}
	\begin{subfigure}[t]{0.32\columnwidth}
		\centering
		\includegraphics[width=\columnwidth]{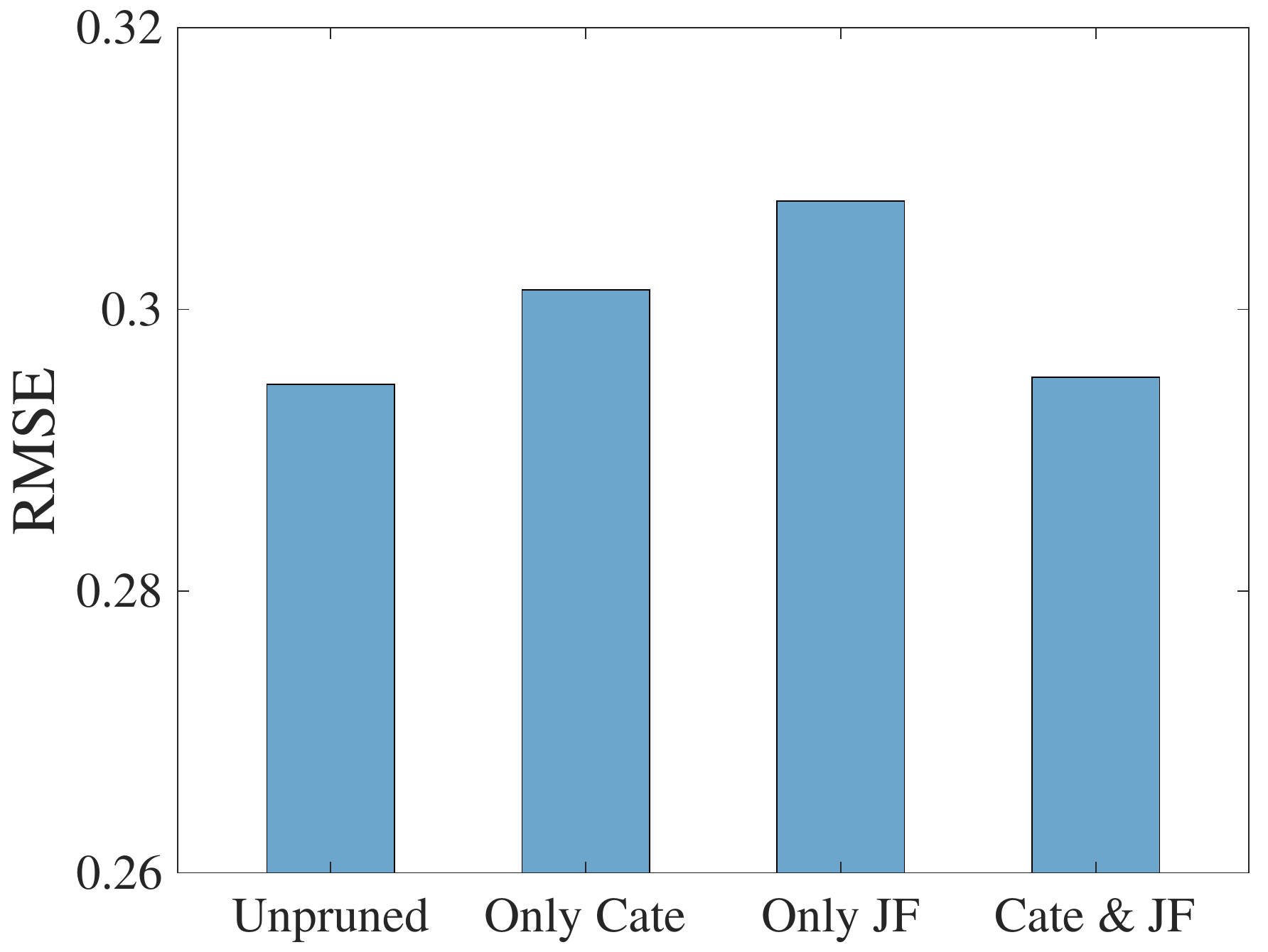}
		\caption{Indiegogo-7K.}\label{fig:lensb}
	\end{subfigure}
	\begin{subfigure}[t]{0.32\columnwidth}
		\centering
		\includegraphics[width=\columnwidth]{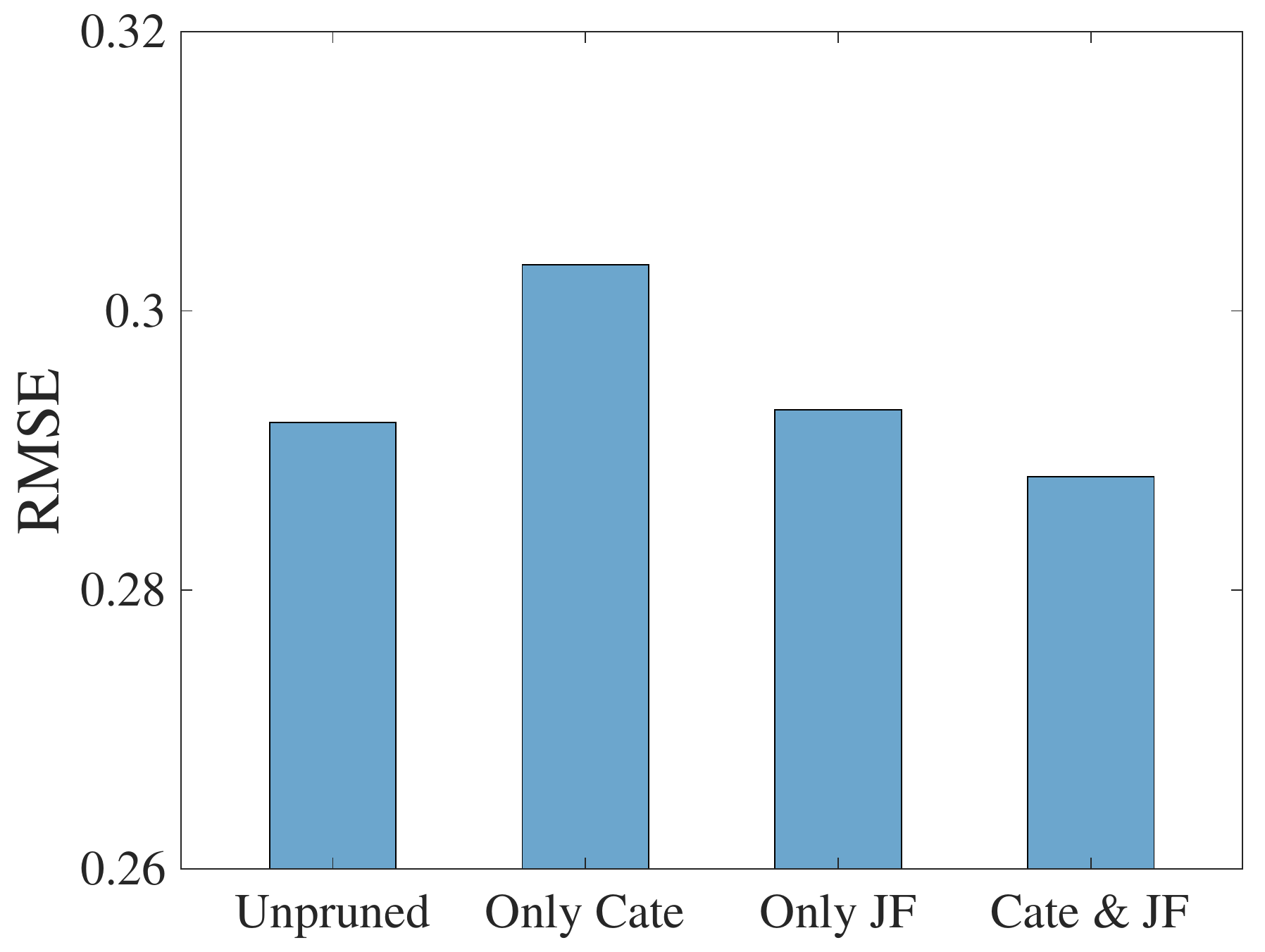}
		\caption{Indiegogo-10K.}\label{fig:lensa}		
	\end{subfigure}
	\begin{subfigure}[t]{0.32\columnwidth}
		\centering
		\includegraphics[width=\columnwidth]{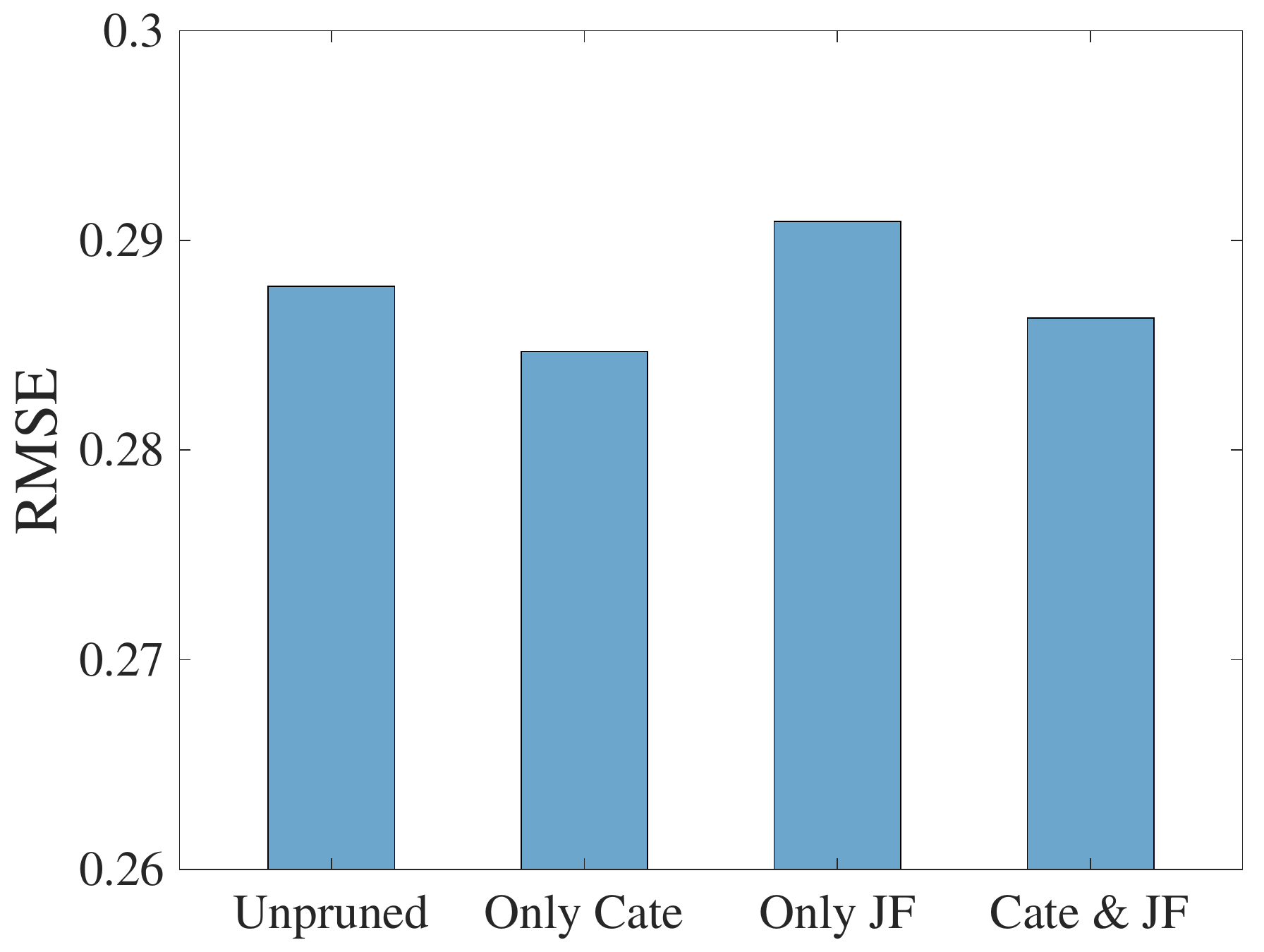}
		\caption{Indiegogo-14K.}\label{fig:lensb}
	\end{subfigure}
	\caption{Influence of pruning approaches.}\label{fig:prun}
\end{figure}

In this section, we evaluate the pruning approach in the PCM module ($\mathcal{T}=24$). Since this approach retains only two physical relationships: projects belong to the same category and on the just-funded section, we test four kinds of variants of GME: (1) \textit{Unpruned} indicates we do not prune. (2) \textit{Only Cate} retains the same category pairs. (3) \textit{Only JF} removes all edges except for those projects in the just-funded section at time $T_g$. (4) \textit{Cate \& JF} method retains the connections of the same category and just-funded section. According to Figure~\ref{fig:prun}, we can find \textit{Cate \& JF} usually promotes the performance best except the RMSE score on Indiegogo-14K. Actually, when we keep the only single physical connection, it decreases the model's performance slightly. After all, it leaves out some needed information. Thanks to \textit{Cate \& JF} avoid this disadvantage to some degree to result in a better prediction performance than \textit{Unpruned}.

\subsubsection{Visualization of the Attention Mechanism.}
\begin{figure}[t]
\centering
\includegraphics[width=0.6\columnwidth]{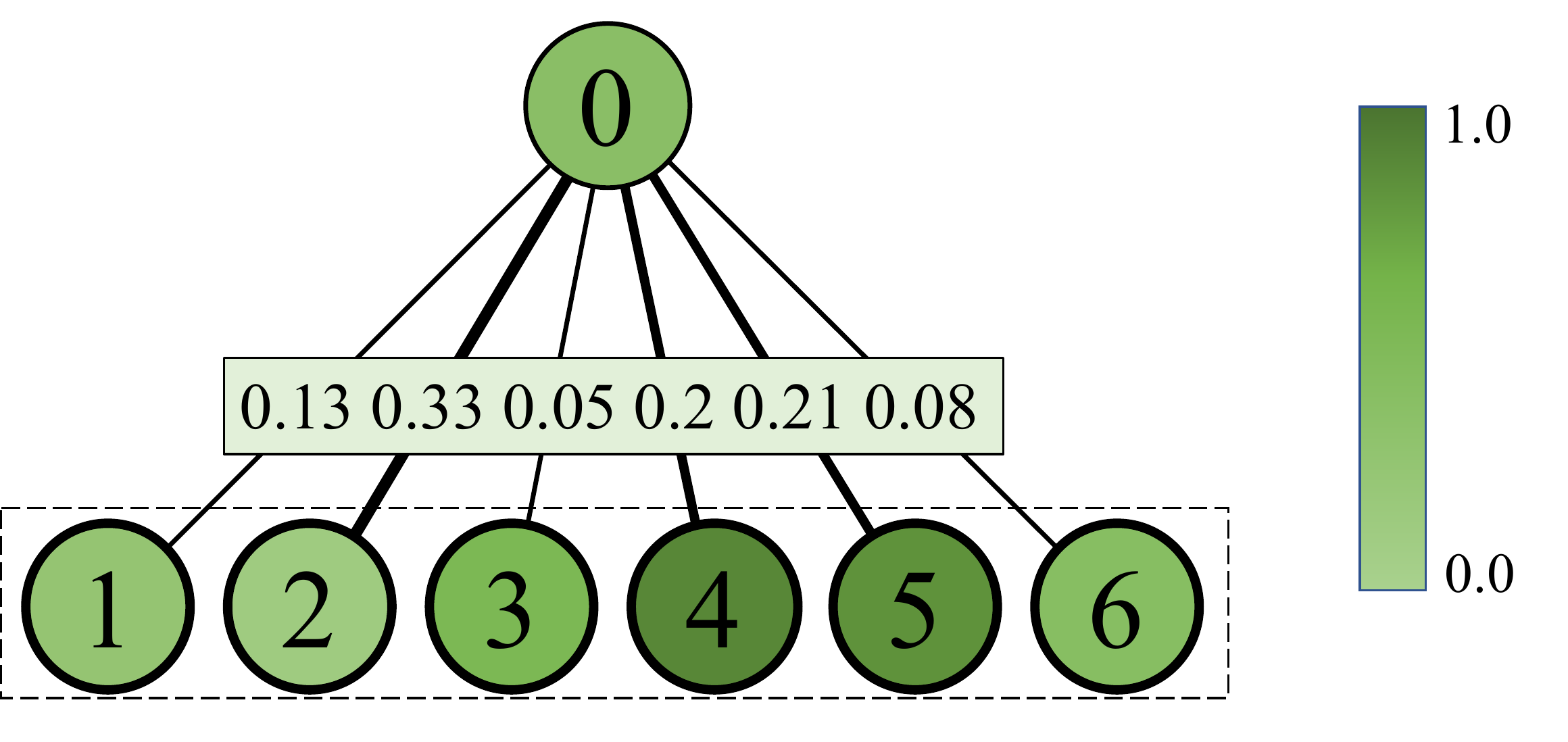}
\caption{Visualization of the Attention Mechanism.}
\label{fig:attentioncase} 
\end{figure}

To further illustrate the learning and expression of the impact distribution from competitors in the PCM module, we visualize the intermediate results of the weight score ${\alpha}_{gi}$ in Eq.7. Table~\ref{table:contents} and Figure~\ref{fig:attentioncase} illustrate the competition relations in a case.

\begin{table}[t]
	\renewcommand\arraystretch{1.1}
	\caption{The contents of sample projects.}\smallskip
	\centering
	\resizebox{0.97\columnwidth}{!}{
		\smallskip\begin{tabular}{ccccc}
			\hline
			\textbf{Id} & \textbf{Project Description} & \textbf{Category} &\textbf{Goal} &\textbf{Duration}\\
			\hline
			0 & Mini PC? Also Notebook PC and Development Board. & Technology
 & 50000 &60\\
 			1 & Invention Designer 3D Printer. & Design & 75000 & 60\\
			2 & WinDMO: A Pocket Sized PC for iOS \& Android. & Technology & 150000 & 60\\
			3 & Danny and the Deep Blue Sea. & Theatre & 7100 & 31\\
			4 & The Shlocker Secure, Convenient Shower Storage! & Design & 30000 & 60\\
			5 & WING : The Accessible Premium Wireless Earphone. & Technology & 18500 & 35\\
			6 & Austin, Texas tiny house rental for \$50 a night. & Community & 48000 & 60\\
			\hline
		\end{tabular}
	}
	\label{table:contents}
\end{table}

In Table~\ref{table:contents}, we list the contents of a set of nodes(projects) from the same batch. In detail, each project is denoted by corresponding Id, Project Description, Category, Declared Pledged Goal, and Declared Funding Duration. Note that, for the Project Description, we only show the projects' titles which include keywords as the limited space. The project with Id 0 is our prediction target. Since there are too many neighboring of a target project, we cannot show all the individuals. So, we randomly select 6 nodes that are numbered 1 to 6 to visualize from the set of neighbor nodes.

Figure~\ref{fig:attentioncase} declares the result generated from our model. We analyze the figure from two perspectives. On the one hand, the numbers in the rectangle box are relation weights. Here we renormalize the score ${\alpha}_{gi}$ to [0, 1], and the sum of all normalized values is 1. In this way, we can easily find that the relation (2-0) has the maximum weight based on project contents. Comparing these relation pairs carefully, project 0 and 2 have a common attention point that they both focus on the development of mini PC. It means that our model can capture meaningful content relevance among different projects. On the other hand, the shade of a node's color indicates corresponding competitiveness quantified by PCM. In detail, we normalize the predicted value (it comes from projecting $\textbf{h}_{i}^{c}$ to 1-dimension value via the MLP layer in the Joint Optimization module) to [0, 1] to visualize the strength of competitiveness. Comparing the shades of color in this figure, project 4 and 5 would also have a great influence on the target project, because their funding strengths are both powerful. 
Therefore, achieving the desired requirements, the phenomenon illustrates that our model is capable of giving consideration to two influences include fund-raising ability and contents of projects at the same time.

\begin{minipage}[t]{\columnwidth}
\centering
 \begin{minipage}{0.52\columnwidth}
  	\makeatletter\def\@captype{table}\makeatother\caption{Performance comparison of two competitiveness quantification methods, $\mathcal{T}=24$.}
	\resizebox{.95\columnwidth}{!}{
		\smallskip\begin{tabular}{ccccc}
			\hline
			\textbf{Metrics} & \textbf{Scale} & \textbf{MLP} & \textbf{Pri-MLP} & \textbf{LSTM} \\
			\hline
			\multirow{3}{*}{MAE}  & 7K & 0.1833 & 0.1802 & \textbf{0.1770} \\
			& 10K & 0.1865 & 0.1815 & \textbf{0.1797} \\
			& 14K & 0.1852 & \textbf{0.1808} & 0.1820 \\
			\hline
			\multirow{3}{*}{RMSE} & 7K & 0.3039 & 0.2976 & \textbf{0.2952} \\
			& 10K & 0.3004 & \textbf{0.2870} & 0.2881 \\
			& 14K & 0.2898 & \textbf{0.2841} & 0.2863 \\
			\hline
		\end{tabular}
	}
	\label{table:mlp_comp}
  \end{minipage}
  \centering
  \begin{minipage}{0.47\columnwidth}
    	\makeatletter\def\@captype{figure}\makeatother
		\centering
		\includegraphics[width=0.77\columnwidth]{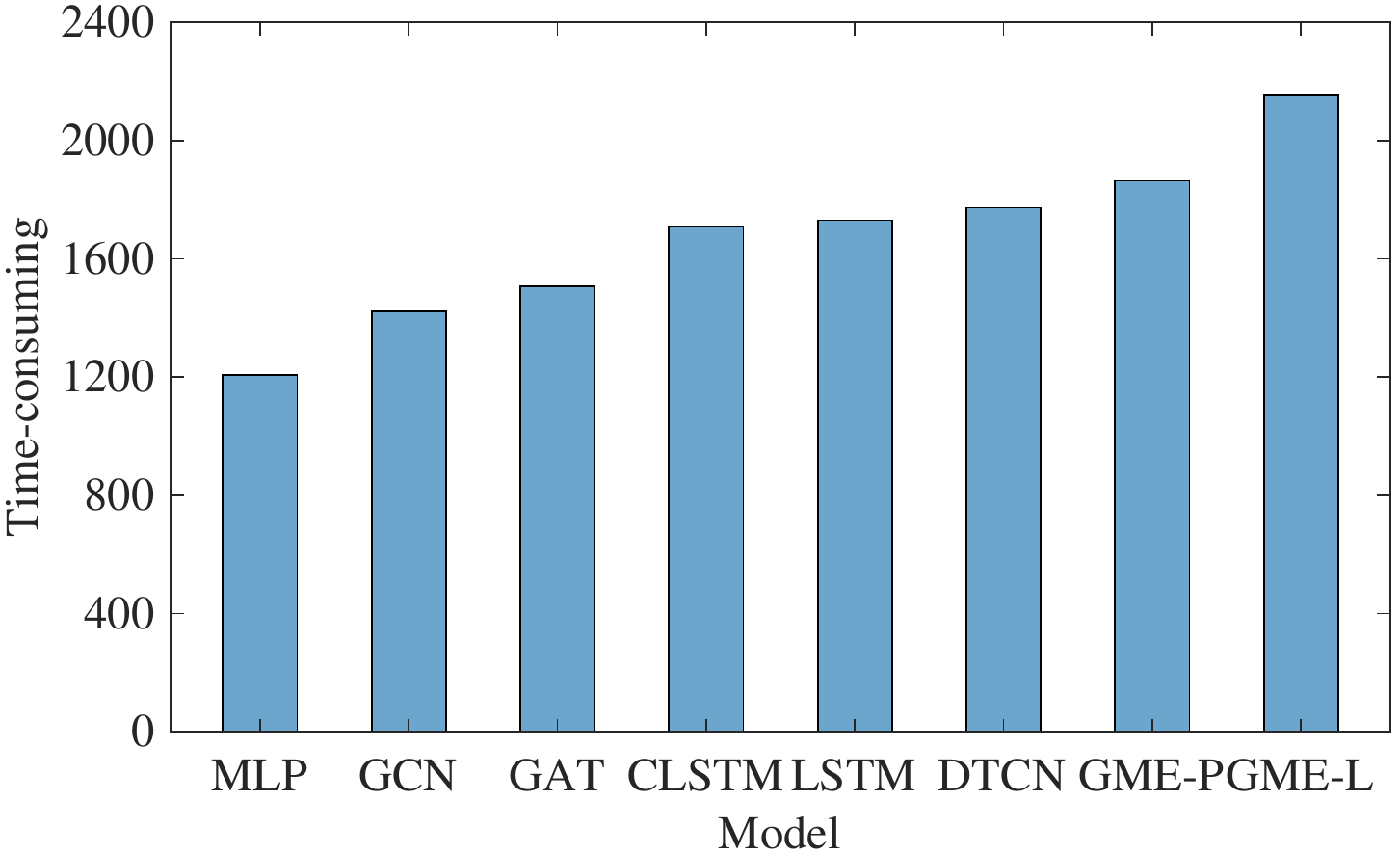}
		\label{fig:timeconsume}
		\caption{Time consuming.}
   \end{minipage}
\end{minipage} 
~\\
 \subsubsection{Comparison of Different Competitiveness Quantification.} 
In section 3.2.1, we propose the MLP-based competitiveness quantification method which adjusts the learning of trends. Compared with LSTM, we aim to achieve good enough prediction accuracy via a more concise approach. Moreover, it is more suitable for large-scale data to replace the LSTM module with the feedforward architecture.
Here we compare the two types to verify our idea.\\

Table~\ref{table:mlp_comp} illustrates the results of the completed GME model with different competitiveness quantification modules. In this section, MLP, Pri-MLP, and LSTM indicate the GME model with different competitiveness predictors: MLP, Priority Knowledge-based MLP, and LSTM, respectively. For the metric of MAE, the performances of LSTM-based GME are relatively better due to its most accurate scores 0.1770 and 0.1797 on two small scale datasets. But Pri-MLP-based GME provides the minimum MAE of 0.1808 on Indiegogo-14K. Focusing on RMSE, Pri-MLP-based GME outperforms the other models that it runs the best races on both Indiegogo-10K and 14K. There is a natural phenomenon that GME with the original MLP predictor should be improved indeed. 
Besides, on the time-consuming level, we record the training time (second) of each deep learning model used in our experiments. As shown in Figure~9, Pri-MLP-based GME (GME-P) converges in a shorter time than LSTM-based GME (GME-L) due to the feedforward competitiveness quantification. Meanwhile, although our proposed models spend more training time, the consumptions of all deep learning models are basically at one amount level. With our promotion degree, this kind of cost is completely acceptable in deep learning research.
According to the analysis above, we find that the simple feedforward network that includes prior knowledge can also play an important role the same as the LSTM module. Certainly, it depends on the particularity of our crowdfunding scene.

\section{Conclusion}

In this paper, we studied the open issue of estimating the fund-raising performance of the unpublished project in the online innovation market. To tackle this task, we proposed an end-to-end Graph-based Market Environment model (GME) to explore the market environment and predict the potential fund-raising performance of the target project. In detail, the innovation market modeling consisted of two modules: project competitiveness modeling and market evolution tracking. Both of them are created as graph-based neural networks, and we trained these two via a joint optimization function. Meanwhile, to enhance the efficiency of our proposed model in the large-scale market environment graph, we designed the MLP-based competitiveness quantification with prior knowledge and the hierarchical updating algorithm. Extensive experiments were conducted on the real-world dataset collected from Indiegogo. The experimental results illustrated the necessity and effectiveness of modeling the market environment to address the fund-raising estimation task. And, GME was able to achieve the best performance amongst most of the experimental settings. 

Certainly, there is also a weakness in our research. Our work is currently limited to the field of online crowdfunding, it has not yet widely used in various scenarios.
But it is not negligible that our framework has great potential to be transferred to deal with some important issues of other applications, such as the cold-start problem in the recommendation system and the internet traffic prediction of the advertisement platform.
In the future, we will explore the commonality between our task and other scenarios to extend our architecture to a more general framework.

\section*{Acknowledgements}
This research was partially supported by grants from the National Key Research and Development Program of China (No. 2016YFB1000904), and the National Natural Science Foundation of China (Grants No. U1605251, 61922073, 61672483). Qi Liu acknowledges the support of the Young Elite Scientist Sponsorship Program of CAST and the Youth Innovation Promotion Association of CAS (No. 2014299).





\bibliography{947-myreference}
\bibliographystyle{elsarticle-num-names} 

%
%
%
\end{document}